\documentclass[10pt,twocolumn,letterpaper]{article}

\usepackage[algorithms]{wacv}      

\usepackage{graphicx}
\usepackage{amsmath}
\usepackage{amssymb}
\usepackage{booktabs}

\usepackage{caption}
\usepackage{subcaption}
\usepackage{wrapfig}
\usepackage{algorithm}
\usepackage{algpseudocode}
\usepackage[pagebackref,breaklinks,colorlinks]{hyperref}

\usepackage[capitalize]{cleveref}
\crefname{section}{Sec.}{Secs.}
\Crefname{section}{Section}{Sections}
\Crefname{table}{Table}{Tables}
\crefname{table}{Tab.}{Tabs.}

\def\wacvPaperID{639} 
\def\confName{WACV}
\def\confYear{2025}
\usepackage{times}
\usepackage{epsfig}
\usepackage{graphicx}
\usepackage{caption}
\usepackage{subcaption}
\usepackage{amsmath}
\usepackage{amssymb}
\usepackage{booktabs}
\usepackage{pifont}
\def\our{GeoGuide}
\def\base{ADM-G}

\newcommand{\cmark}{\ding{51}}%
\newcommand{\xmark}{\ding{55}}%

\def\e{\varepsilon{}}
\def\N{\mathcal{N}}
\def\E{\mathbb{E}}
\def\cost{\mathcal{L}}

\def\R{\mathbb{R}}

\begin{document}

\title{\our{}: Geometric guidance of diffusion models}

\author{
Mateusz Poleski\\
Jagiellonian University, Faculty of Mathematics and Computer Science\\
{\tt\small mateusz.poleski@student.uj.edu.pl}
\and
Jacek Tabor\\
Jagiellonian University, Faculty of Mathematics and Computer Science\\
{\tt\small jacek.tabor@uj.edu.pl}
\and
Przemysław Spurek\\
Jagiellonian University, Faculty of Mathematics and Computer Science\\
IDEAS NCBR\\
{\tt\small przemyslaw.spurek@uj.edu.pl}
}

\maketitle
\thispagestyle{empty}



\begin{abstract}
Diffusion models are among the most effective methods for image generation. This is in particular because, unlike GANs, they can be easily conditioned during training to produce elements with desired class or properties. 

However, guiding a pre-trained diffusion model to generate elements from previously unlabeled data is significantly more challenging. One of the possible solutions was given by the ADM-G guiding approach. Although ADM-G successfully generates elements from the given class, there is a significant quality gap compared to a model originally conditioned on this class. In particular, the FID score obtained by the ADM-G-guided diffusion model is nearly three times lower than the class-conditioned guidance. We demonstrate that this issue is partly due to ADM-G providing minimal guidance during the final stage of the denoising process. To address this problem, we propose \our{}, a guidance model based on tracing the distance of the diffusion model's trajectory from the data manifold.
The main idea of \our{} is to produce normalized adjustments during the backward denoising process. As shown in the experiments, \our{} surpasses the probabilistic approach ADM-G with respect to both the FID scores and the quality of the generated images.
\end{abstract}


\section{Introduction}

Diffusion models are crucial for generating images with specified characteristics. Compared to GAN models, their benefit is that they can be easily conditioned to produce outcomes with the desired attributes \cite{dhariwal2021diffusion,rombach2022high,croitoru2023diffusion}.

\begin{figure}[]
    \centering
    \!\begin{tabular}{@{}c@{}c@{}}
        \rotatebox{90}{ \small \quad \our{} \qquad \base{}} &
        \includegraphics[width = 0.485\textwidth]{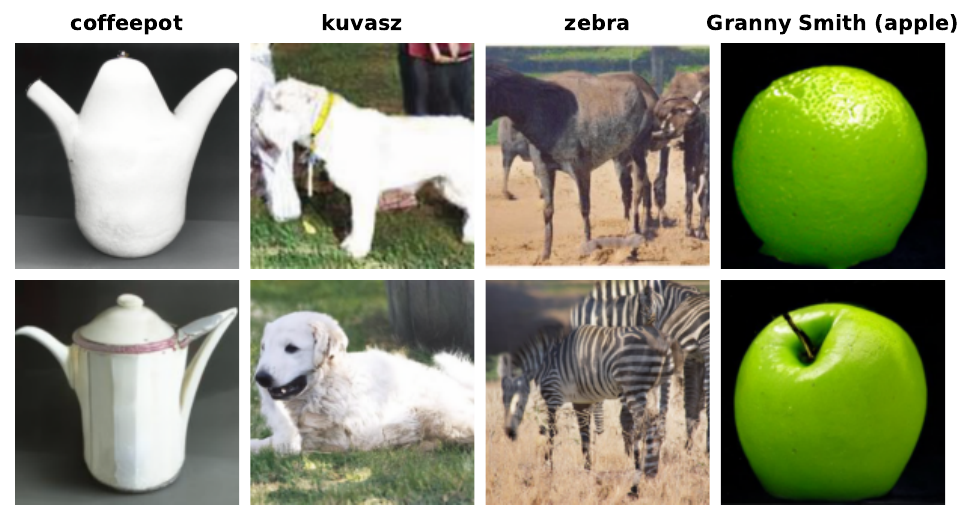} \\
    \end{tabular}
    \caption{In our paper, we propose a new approach to guidance of diffusion models, called \our{}. In contrast to \base{}~\cite{dhariwal2021diffusion} we use updates with the same norm and consequently keep the guided diffusion process close to the manifold of a given class. Observe that this allows us to construct images with more details characteristic of a given class, resulting in a decrease in the FID score from 12 in \base{} to 7.32 in \our{}, see Table \ref{tab:our_resuls_comparison}. The images were constructed with the same diffusion noise for \base{} and \our{}.
    }
    \label{fig:samples_comparison_uncond_teaser}
\end{figure}

However, given a pre-trained diffusion model, it is not trivial to construct images that belong to the class that was not considered earlier in the conditioning process. One of the possible solutions is given by the guidance model \base{} \cite{dhariwal2021diffusion}. Roughly speaking, to produce an image from a new class, we first train a classifier for this class and then add the rescaled gradient during the backward process.
\base{} introduces guidance through probabilistic principles, for a more detailed description, see Section~\ref{cl_giud}.
Unfortunately, \base{} shows a notable difference in the FID score between the model that only uses guided techniques and the model with class-conditioned guidance. Specifically, the ADM-G model achieves a 4.59 FID score with class-conditioned guidance, while the guided model obtains only 12 FID score, as referenced in Table 4 of \cite{dhariwal2021diffusion}.

\begin{figure}[]
    \centering
    \includegraphics[width = 0.45\textwidth]{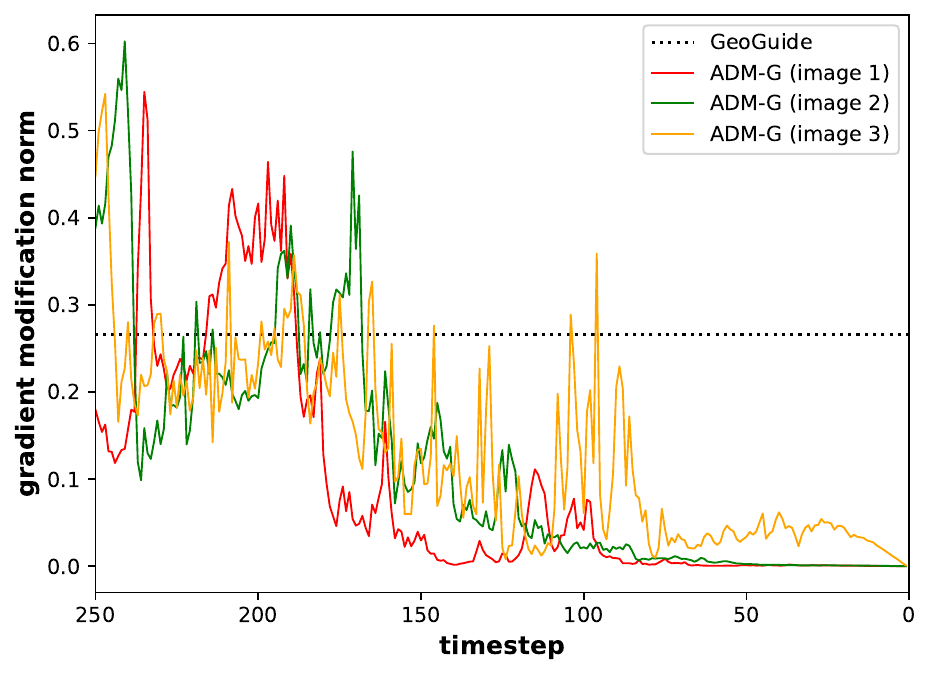}
    \caption{Norm values of the gradient modification factor applied at each iteration of the classifier guided diffusion sampling backward process. Comparison of image generated with \our{} and three random images generated with \base{}. Observe that in the case of the vanilla guidance (probabilistic approach) the norm of the modification at the last steps of the diffusion process is close to zero, which results in less detail in the produced images, see Figure~\ref{fig:samples_comparison_uncond_teaser}. 
    }
    \label{fig:modification-norm}
\end{figure}

This paper aims to reduce the gap between the quality of images generated by diffusion models guided on previously unlabeled data compared to class-conditioned guides, see Figure \ref{fig:samples_comparison_uncond_teaser}.
To do so, we switch perspectives from probabilistic to metric. We postulate that the trajectories of the guided model should be close to the metric properties of the unguided ones. 
It occurs that by applying \base{} guidance, the norms of adjustment in to the backward diffusion denoising decrease significantly in the last iterations of the denoising procedure, see Figure~\ref{fig:modification-norm}.
As a result, the images become nearly unguided at half of their trajectory and consequently lose the ability to produce details specific to the given class, see Figure \ref{fig:samples_comparison_uncond_teaser} and Figure \ref{fig:cut_guidance}.

To address this issue, we shift the perspective from a probabilistic framework to a metric-based method. We theoretically demonstrate that the trajectory of the diffusion model lies close to the data manifold. Using such intuition, we propose a new guidance model \our{}\footnote{Code is available at \url{https://github.com/mateuszpoleski/geoguide}}, which uses fixed length updates to force the denoising process to be as close as possible to the data manifold. \our{} use norm of classification gradient to control updates. Therefore, our model is easy to implement and outperforms the probabilistic approach with respect to FID score and the quality of generated images.

Concluding, the main contributions of the paper are the following:
\begin{itemize}
\vspace{-0.2cm}
    \item we propose a new guidance model \our{}, motivated by the metric properties of the trajectories of the diffusion process,
\vspace{-0.2cm}    
    \item \our{} is easy to implement and controls the norm of the guidance,
\vspace{-0.2cm}    
    \item \our{} outperforms \base{} in pure guidance with respect to FID score and quality of generated images.
\vspace{-0.3cm}    
\end{itemize}

\section{Related Works}

\label{rob_classifier_descr}
Our method aims to improve sample generation of pre-trained diffusion models \cite{dhariwal2021diffusion,zhang2023adding,yang2023diffusion}. It can be easily incorporated into existing models as it does not modify the structure or inner workings of the network\cite{luo2024diff,deja2022analyzing}. It only modifies sampling logic, which allows one to easily leverage the power of already existing diffusion models and classifiers. In \cite{kawar2023enhancingdiffusionbasedimagesynthesis} authors present another approach to enhance sampling quality in diffusion models. They defined alternative way of training for time-dependent adversarially robust classifier, and use it as guidance for a generative diffusion model. These classifier gradients are better aligned with human perception, and could better guide a generative process towards semantically meaningful images. In Section~\ref{rob_classifier} we tried to combine robust classifier with our \our{} and evaluated results.

Another interesting technique for improving guidance in score-based diffusion models is Discriminator Guidance \cite{kim2023refininggenerativeprocessdiscriminator}. The authors propose integrating a discriminator, commonly used in Generative Adversarial Networks (GANs), to guide the generative process. The key idea is to use the discriminator to evaluate and refine the intermediate states of the generative process, improving overall sample quality. Integrating discriminator guidance helps mitigate mode collapse and improves sample diversity and fidelity. This novel approach combines discriminator and vanilla classifier guidance in the generation process. It is possible that integrating \our{} into this approach, instead of a vanilla guidance, could lead to even better results.

As we can see in Figure \ref{fig:distribution_comparison}, diffusion models, especially when used with high guidance scale values to achieve optimal image quality, are prone to limited output diversity. One potential solution to this problem is the Condition-Annealed Diffusion Sampler (CADS) \cite{sadat2024cadsunleashingdiversitydiffusion}. In this approach, the sampling strategy anneals the conditioning signal by adding scheduled, monotonically decreasing Gaussian noise to the conditioning vector during inference to balance diversity and condition alignment. This results in increased generation diversity, especially at high guidance scales, with minimal loss of sample quality. This method has shown good results in the context of classifier-free guidance. However, it might also be possible to successfully incorporate it into \our{}.

\section{Diffusion models}

Diffusion models are generative algorithms that produce new samples using an iterative denoising procedure. We start from Gaussian noise $x_T$ and gradually produce less noisy samples $x_{T-1},x_{T-1}, \ldots,x_{0}$. Ultimately, $x_{0}$ comes from the data manifold. In each time step $t$, we have a certain noise level, and $x_t$ is a mixture of signal $x_{0}$ and Gaussian noise $\e$. The time step $t$ determines the level of noise. 
Diffusion models are trained using random elements and time steps to produce denoised $x_{t-1}$ from $x_t$. This process is usually modeled by U-Net~\cite{ho2020denoising}.

Diffusion models use two processes: the forward and reverse diffusion process.
The first is simple. Let $q(x_0)$ denote the data distribution $x_0 \sim q(x_0)$.
The forward diffusion process adds a small amount of Gaussian noise to the sample in
$T$ steps, producing a sequence $x_{0}, \ldots, x_{T}$. Such process is controlled by $\{ \beta_{t} \in (0,1) \}_{t=1}^T$:
\begin{equation} \label{eq:0}
q(x_t|x_{t-1}) := \N(x_t ; \sqrt{1-\beta_t} x_{t-1},\beta_{t}I).
\end{equation}
Such formula allows to obtain $x_t \sim q(x_t|x_0)$ in one step instead apply repeatedly $q$:
\begin{equation}
\label{eq:epsilon}
\begin{split}
q(x_t|x_0) &  = \N(x_t ; \sqrt{ \bar \alpha_t } x_0, (1-\bar \alpha_t) I) \\
& = \sqrt{ \bar \alpha_t } x_0 + \e \sqrt{ 1-\bar \alpha_t }, \mbox{ } \e \sim \N(0,I),
\end{split}
\end{equation}
where  $\alpha_t = 1-\beta_t$ and $\bar \alpha_t = \prod_{s=0}^t \alpha_t$.

Typically we choose the schedule $\beta_t$ in such a way that $\bar \alpha_T$ is close to $0$, which implies that $q(x_T|x_0)$ is close to the distribution $\N(0,I)$. A simplest schedule for $\bar \alpha_t$ which satisfies this condition can be given by
\begin{equation}
    \bar \alpha_t=1-t/T.
\end{equation}
Additionally, we usually assume that the $\beta_t$ is increasing so that we have more denoising steps at the end of the backward process, which yields better quality of generated images (while we simultaneously allow larger denoising steps at the beginning of the backward process) \cite{ho2020denoising,nichol2021improved}.  

By applying Bayes' theorem, it can be determined that the posterior distribution $q(x_{t-1}|x_t,x_0)$ is a Gaussian, characterized by the mean $\tilde \mu_t (x_t, x_0)$ and the variance $\tilde \beta_t$ as specified below:
$$
q(x_{t-1}|x_t,x_0) = \N( x_{t-1}; \tilde \mu_t(x_t,x_0),\tilde \beta_t I ),
$$
where 
$\tilde \mu_t(x_t,x_0) := \frac{\sqrt{\bar \alpha_{t-1}}\beta_t}{1- \bar \alpha_t} x_0 + \frac{\sqrt{\alpha_t}(1-\bar \alpha_{t-1})}{1-\bar \alpha_{t}}x_t $ and $\tilde \beta_t := \frac{1-\bar \alpha_{t-1}}{1-\bar \alpha_{t}} \beta_t$.

Theoretically, we can draw samples from the data distribution $q(x_0)$. We begin by sampling from $q(x_T)$ and then proceed by sampling the reverse steps $q(x_{t-1}| x_t)$ until we arrive at $x_0$. With appropriate choices for $\beta_t$ and $T$, the distribution $q(x_T)$ approximates an isotropic Gaussian distribution, making the sampling of $x_T$ straightforward.

Since the data distribution is unknown, we use a neural network to approximate $q(x_{t-1}|x_t)$. In the reverse diffusion process, we approximate these conditional probabilities. In \cite{sohl2015deep} the authors show that $q(x_{t-1}|x_t)$  approaches a diagonal Gaussian distribution as $T \to \infty$ and, correspondingly, $\beta_t \to 0$. In the reverse diffusion process, we train a
neural network to predict a mean $\mu_{\theta}$ and a 
diagonal covariance matrix $\gamma_t I$:
$$
p(x_{t-1}|x_{t}):= \N(x_{t-1};\mu_{\theta}(x_t,t), \gamma_t I).
$$
To ensure that $p(x_0)$ captures the actual data distribution $q(x_0)$, we can optimize the corresponding variational lower bound. Such cost function is theretialy motivated, but in practice \cite{ho2020denoising} propose to  do not directly parameterize $\mu_{\theta}(x_t,t)$ as a neural
network, but instead train a model $\e_{\theta}(x_t, t)$ to predict $\e$ from equation (\ref{eq:epsilon}).

The following outlines the simplified objective:
$$
\cost{}:= \E_{t\sim [1,T],x_0 \sim q(x_0), \e \sim \N (0,I)} \left[  \| \e - \e_{\theta}(x_t,t) \|^2 \right]
$$
During sampling, we can use substitution to derive $\mu_{\theta}(x_t, t)$ from $\e_{\theta}(x_t, t)$:
$$
\mu_{\theta}(x_t,t) = \frac{1}{\sqrt{\alpha_t}} \left( x_t - \frac{1-\alpha_t}{\sqrt{ 1-\bar \alpha_t }} \e_{\theta}(x_t,t)  \right).
$$

It is important to note that $\cost{}$ does not offer any learning signal for $\gamma_t$. According to \cite{ho2020denoising}, rather than learning $\gamma_t$, it can be set to a constant value, either $\beta_t$ or $\tilde \beta_t$. These constants represent the upper and lower limits for the actual reverse-step variance.

\section{Classifier guidance and \our{}}

This section presents our geometric approach to guidance in diffusion models. We start with the general idea behind the guidance of diffusion models, and then we proceed with the description of \base{} and \our{}.

Let us first recall that given a controlling variance schedule $\beta_t \in (0,1)$, in the forward process, we start from the point $x_0$ in the data manifold $M$, and put 
$$
x_t=\sqrt{1-\beta_t}x_{t-1}+\sqrt{\beta_t} \varepsilon_t, \text{ where }\varepsilon_t \sim \N(0,I).
$$
Finally, we return $x_T$.
In the backward pass, we start with randomly chosen $x_T \sim \N(0,I)$, and put
$$
x_{t-1}=\mu(t,x_t)+\gamma_t \varepsilon_t,
$$
where $\gamma_t$ are constants, $\mu$ is a deep network (typically given by U-NET) and $\varepsilon_t \sim \N(0,I)$.
The function $\mu$ and the constants $\gamma_t$ are chosen so that the trajectories of the forward and backward pass constructs cannot be distinguished.

The task of guidance lies in generating data from a pre-trained diffusion model from a given class $y$. To do so, we usually train a classifier on $p(y|x)$ (optimally also on elements from $y$ with added noise), and adjust the backward trajectory by the rescaled gradient of the classifier:
$$
x_{t-1}=\mu(t,x_t)+\sqrt{\gamma_t} \varepsilon_t+
s \cdot A(p(y|x),\nabla p(y|x)),
$$
where $s$ is the scaling parameter.
The problem lies in choosing a function $A$ that would lead to the optimal generation of points of class $y$. In the case of \base{} (see Algorithm~\ref{alg:vanilla}) we have
$$
A=\gamma_t \nabla \log p(y|x)=\frac{\gamma_t}{p(y|x)} \cdot \nabla p(y|x),
$$
while in the case of our model \our{} (see Algorithm~\ref{alg:our}) we have
$$
A=\frac{\sqrt{D}}{T} \cdot \frac{\nabla p(y|x)}{\|\nabla p(y|x)\|}.
$$
Before presenting the justifications for both \base{} and \our{}, observe that the complexity of both adjustments is similar.

\begin{figure*}
\centering
\begin{subfigure}{.50\textwidth}
    \begin{tabular}{@{}c@{}c@{}}
        \rotatebox{90}{ \,\, \quad $100\%$ \quad \qquad $30\%$} &
        \includegraphics[width = 0.9\textwidth]{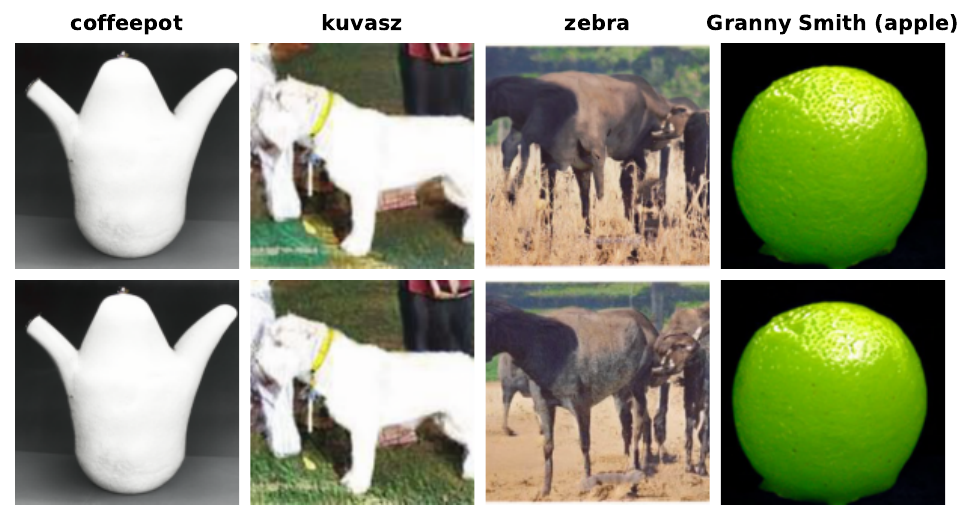} \\
    \end{tabular}
    \centering
    \caption{\base{}}
    \label{fig:cut_guidance_base}
\end{subfigure}%
\begin{subfigure}{.50\textwidth}
    \begin{tabular}{@{}c@{}c@{}}
        \rotatebox{90}{ \,\, \quad $100\%$ \quad \qquad $30\%$} &
        \includegraphics[width = 0.9\textwidth]{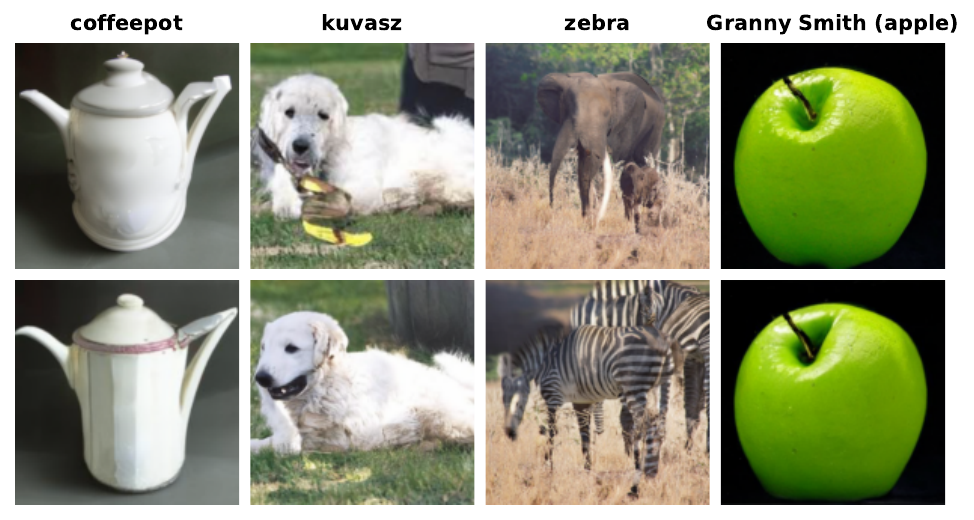} \\
    \end{tabular}
    \centering
    \caption{\our{}}
    \label{fig:cut_guidance_our}
\end{subfigure}
\caption{Comparison of results when guidance is turned off after first 30\% of iterations vs fully guided samples. \base{} is not effective during last 70\% of iterations, whereas \our{} is still significantly improving quality of generated images.}
\label{fig:cut_guidance}
\end{figure*}

\paragraph{Classifier Guidance in \base{}}\label{cl_giud}
A key characteristic of diffusion models is their ability to generate elements based on arbitrary classes. A classifier $p(y|x)$ can enhance a diffusion generator. As demonstrated in \cite{sohl2015deep,song2020score,dhariwal2021diffusion}, a pre-trained diffusion model can be conditioned using classifier gradients. Specifically, a classifier $p_{\phi}(y|x_t, t)$ can be trained on noisy images $x_t$, and the gradients $ 
\nabla_{x_t} \log p_{\phi}(y|x_t, t)$ can then be used to steer the diffusion sampling process towards a specific class label $y$.
For simplicity, we adopt the notation $p_{\phi}(y|x_t, t) = p_{\phi}(y|x_t)$ and $\e_{\theta}(x_t, t) = \e_{\theta}(x_t)$, acknowledging that these represent distinct functions for each time step $t$ and that during training, the models must be conditioned on the time step $t$.

Let us assume that we have an unconditional reverse noising process $p_{\theta}(x_t|x_{t+1})$. To incorporate the label $y$ into the process, we can sample each transition as follows: 
$$
p_{\theta,\phi}(x_t|x_{t+1}, y) = Z p_{\theta}(x_t|x_{t+1})p_{\phi}(y|x_t), 
$$
where $Z$ serves as a normalizing constant \cite{dhariwal2021diffusion}.
In practical applications \cite{song2020score}, this can be approximated by a perturbed Gaussian distribution. In this section, we will revisit this derivation follow \cite{dhariwal2021diffusion}.

Note that our diffusion model estimates the prior time step $x_t$ from the subsequent time step $x_{t+1}$ utilizing a Gaussian distribution:
\begin{equation}
\begin{split}
p_{\theta} (x_t|x_{t+1}) & =  \N(\mu(t+1,x_{t+1}), \Sigma),\\
\log p_{\theta} (x_t|x_{t+1}) & =  -\frac{1}{2}(x_t - \mu )^T \Sigma^{-1} (x_t - \mu ) + C,
\end{split}
\end{equation}
where $\Sigma=\gamma_{t+1}I$.

It can be assumed that $\log p_{\phi}(y|x_t)$ exhibits low curvature relative to $\Sigma^{-1}$. This assumption holds true in the scenario of an infinite number of diffusion steps, where $\| \Sigma \| \to 0 $. Under these conditions, log $p_{\phi}(y|x_t)$ can be approximated by performing a Taylor expansion around $x_t = \mu$ as:
\begin{equation}
\begin{split}
&\log p_{\phi}(y|x_t)  =\\
 & \approx \log p_{\phi}(y|x_{t})|_{x_t=\mu} + (x_t-\mu) \nabla_{x_t} \log p_{\phi} (y|x_t)|_{x_t=\mu}\\
& = (x_t-\mu) g +C_1,
\end{split}
\end{equation}
where $g = \nabla_{x_t} \log p_{\phi} (y|x_t)|_{x_t=\mu} $, and $C_1$is a constant. 
Therefore
\begin{equation}
\begin{split}
&\log( p_{\theta} (x_t | x_{t-1} ) p_{\theta}(y|x_t) )  \\
&\approx - \frac{1}{2} (x_t - \mu)^T\Sigma^{-1}(x_t - \mu)+ (x_t - \mu)g+C_2 \\
&= \log p(z) + C_4 \mbox{ }, z \sim \N(\mu + \Sigma g, \Sigma) 
\end{split}
\end{equation}

The constant term $C_4$ can be disregarded, as it includes in the normalizing coefficient $Z$. Consequently, we have determined that the conditional transition operator can be approximated by a Gaussian, akin to the unconditional transition operator, but with its mean adjusted by $\Sigma$. Algorithm~\ref{alg:vanilla} outlines the related sampling algorithm. In \cite{dhariwal2021diffusion}, the authors introduce an optional scale factor $s$ for the gradients.


\begin{algorithm}[!t]
    \caption{ADM-G Classifier guided diffusion sampling, given a diffusion model $\mu(t,x_t), \gamma_t$, classifier $p(y|x_t)$, and gradient scale $s$.
    \label{alg:vanilla}}
    \begin{algorithmic}
        \State {\bf Input:} class label $y$, gradient scale $s$
        \State $x_T \leftarrow$ sample from $N (0, I)$
    \For{$t \leftarrow T$ to $1$}
        \State $\varepsilon_t$ sample from $\N(0,I)$
        \State $A_t=\gamma_t \nabla \log p(y|x_t)$
        \State $x_{t-1}=\mu(t,x_{t})+\sqrt{\gamma_t}\varepsilon_t + s A_t $
    \EndFor
    \State {\bf return} $x_{0}$    
    \end{algorithmic}
    \end{algorithm}

\begin{algorithm}[!t]
    \caption{\our{}: Classifier guided diffusion sampling, given a diffusion model $\mu(t,x_t), \gamma_t$, classifier $p(y|x_t)$, and gradient scale $s$.
    \label{alg:our}}
    \begin{algorithmic}
        \State {\bf Input:} class label $y$, gradient scale $s$
        \State $x_T \leftarrow$ sample from $N (0, I)$
    \For{$t \leftarrow T$ to $1$}
        \State $\varepsilon_t$ sample from $\N(0,I)$
        \State $A_t=\frac{\sqrt{D}}{T}\frac{\nabla p(y|x_t) }{ \|\nabla p(y|x_t)\| }$
        \State $x_{t-1}=\mu(t,x_{t})+\sqrt{\gamma_t}\varepsilon_t + s A_t$
    \EndFor
    \State {\bf return} $x_{0}$    
    \end{algorithmic}
    \end{algorithm}

\paragraph{Motivation of \our{}}
This part introduces \our{}, which takes advantage of the metric properties of the underlying space rather than depending on probability theory.
We can interpret the forward diffusion process as a stochastic process that starts with the data manifold $M \subset \R^D$ and ends in the distribution $\N(0,I)$. In the backward process, we try to emulate the behavior of the forward process by reversing the time direction.

The forward diffusion process adds a small amount of Gaussian noise to the rescaled sample in
$T$ steps, producing a sequence $x_{0}, \ldots, x_{T}$. Using \eqref{eq:epsilon} we know that
$$
x_t=\sqrt{\bar \alpha_t}x_0+\sqrt{1-\bar \alpha_t}\varepsilon \in 
\sqrt{\bar \alpha_t}M+\varepsilon
, \text{ where }\varepsilon \sim \N(0,I).
$$
Consequently, 
$$
d(x_t;\sqrt{\bar \alpha_t}M) \leq
d(x_t,\sqrt{\bar \alpha_t}x_0)=
\sqrt{1-\bar \alpha_t}\|\varepsilon\|.
$$
Since in fact we only add noise to the element $x_0$, the closest element to $\sqrt{\bar \alpha_t}x_0+\sqrt{1-\bar \alpha_t}\varepsilon$ from $\sqrt{\bar \alpha_t}M$ would typically be $\sqrt{\bar \alpha_t}x_0$.
Thus, we obtain the approximation
$$
d(x_t;\sqrt{\bar \alpha_t}M) \approx
\sqrt{1-\bar \alpha_t}\|\varepsilon\|.
$$
Since $\varepsilon \sim \N(0,I)$, and the dimension $D$ of the space is large, by the law of large numbers, we obtain $\|\varepsilon\| \approx \sqrt{D}$.
In conclusion, we see that the distance of the flow of the diffusion process from the (rescaled) data manifold is given by
\begin{equation} \label{eq}
d(x_t;\sqrt{\bar \alpha_t}M) \approx
\sqrt{1-\bar \alpha_t} \cdot \sqrt{D}.
\end{equation}

Since the diffusion process satisfies the metric criterion described above, our intuition suggests that any perturbations to achieve the desired changes should be implemented without changing this criterion. Thus, modifications that consider only the gradient are sub-optimal, as the value of the gradient of the classifier is inconsistent on the trajectory, see Figure~\ref{fig:modification-norm}.
Consequently, we postulate that the norm of perturbation of the backward process should be consistent throughout the backward process to influence the trajectory evenly throughout the backward process.

Consequently, the main idea behind \our{} 
lies in the observation, that if during denoising we guide the model by the adjustment with norm proportionally small to the above distance, the trajectories would still satisfy \eqref{eq}. 



\begin{figure*}[!h]
    \centering
    \begin{tabular}{@{}c@{}c@{}}
        \rotatebox{90}{ \, \quad \our{} \qquad \base{}} &
        \includegraphics[width = 0.9\textwidth]{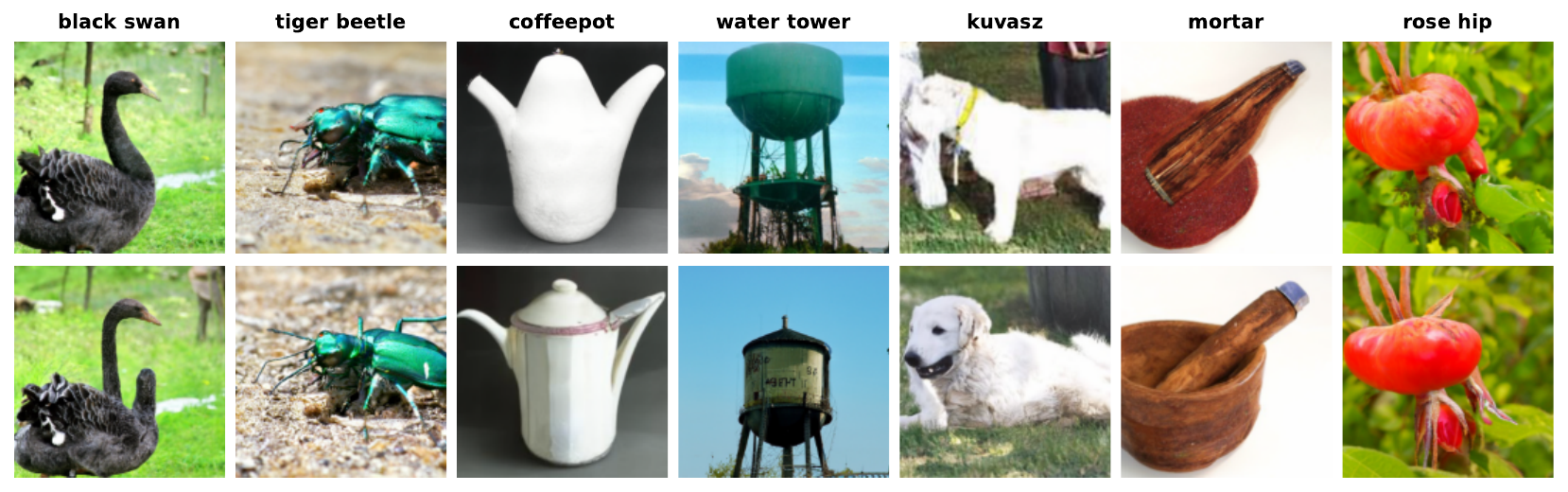} \\
        \rotatebox{90}{ \, \quad \our{} \qquad \base{}} &
        \includegraphics[width = 0.9\textwidth]{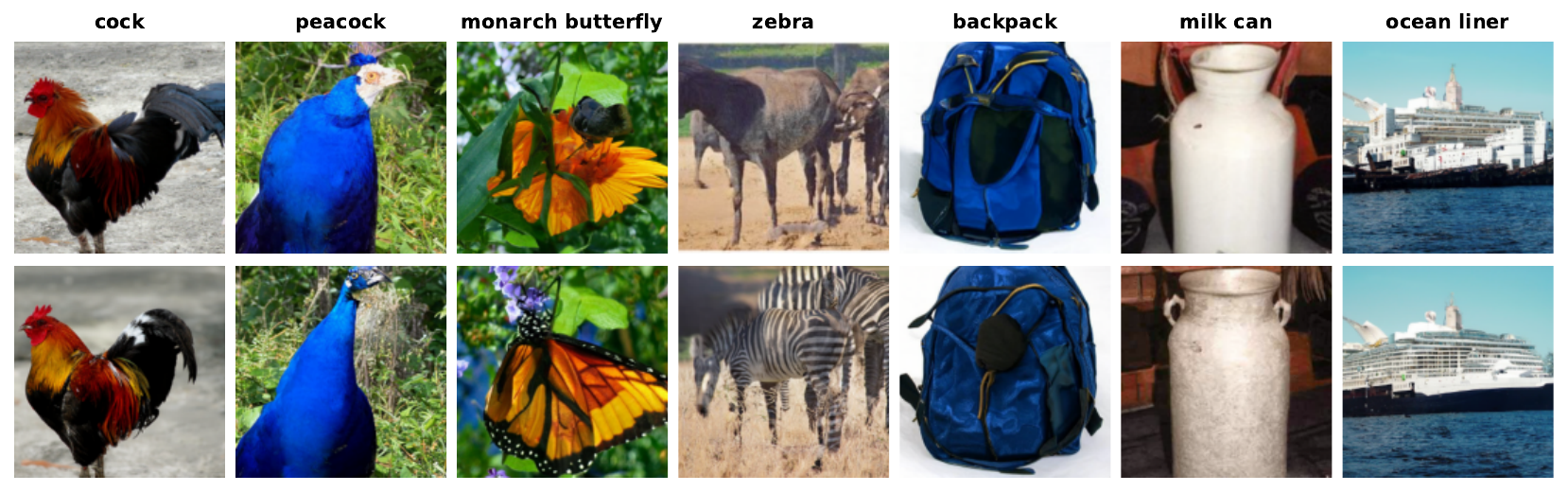} \\
        \rotatebox{90}{ \, \quad \our{} \qquad \base{}} &
        \includegraphics[width = 0.9\textwidth]{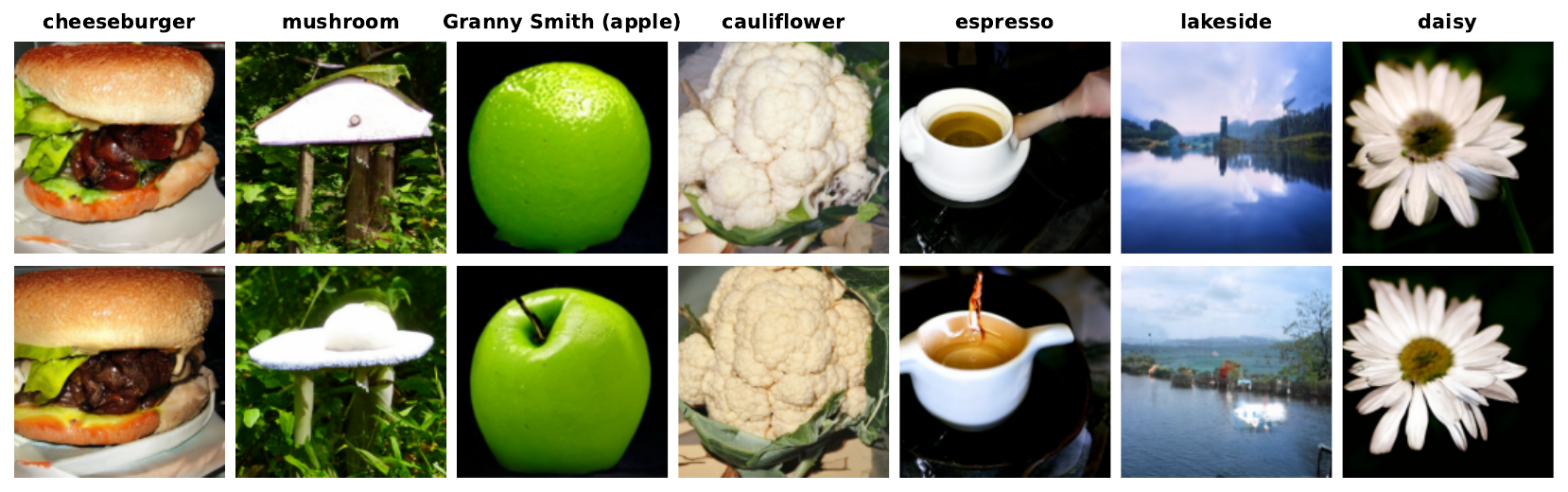}\\
    \end{tabular}
    \caption{Images generated by guided diffusion using the same noise (random seed) and class label, with a vanilla \cite{dhariwal2021diffusion} (FID 12.00, top) and a geometric (FID 7.32, bottom) guidance. Observe that images generated by \our{} are typically much more detailed. In our opinion, this is because the role of the classifier gradient is also important at the end of the backward process. In the \base{}, the norm of the modification at the last steps of the process is close to zero, while in \our{} it stays relevant during the entire process, see Figure~\ref{fig:modification-norm} }
    \label{fig:samples_comparison_uncond}
\end{figure*}



Thus assume that we have a deterministic function $v(x)$ by applying which we would like to modify the trajectory of the backward process, where by the default we may think of $v(x)=\nabla p(y|x)$. Then we would normalize the norm of $v$ to make it proportional to $\sqrt{D(1-\bar \alpha_t)}$, which since $\bar \alpha_T\approx 1$, for large $t$ is close to $\sqrt{D}$. On the other hand, since this perturbation is deterministic, we would also normalize it by the number of steps $T$ in the diffusion process. 
Finally, the adjustment will be given by
\begin{equation} \label{eq:var}
A_t=\frac{\sqrt{D}}{T}\sqrt{1-\bar \alpha_t} \frac{v(x)}{\|v(x)\|}.    
\end{equation}
Such a strategy could be easily implemented, see the Table \ref{tab}, however it still could be improved by taking into account properties of the guidance given by the gradient of the classifier.

\begin{table*}[]
\centering
\begin{tabular}{ccccccccc}
\hline 
Model & Conditional & Guidance & Scale & FID & sFID & IS & Precision & Recall \\
\hline 
ADM & \xmark & \xmark & & 26.21 & $\mathbf{6.35}$ & 39.70 & 0.61 & 0.63 \\
\base{} & \xmark & \cmark & 1.0 & 33.03 & 6.99 & 32.92 & 0.56 & $\mathbf{0.65}$ \\
\base{} & \xmark & \cmark & 10.0 & 12.00 & 10.40 & 95.41 & 0.76 & 0.44 \\
\our{} & \xmark & \cmark & 0.15 & $\mathbf{7.32}$ & 7.98 & $\mathbf{243.34}$ & $\mathbf{0.77}$ & 0.42 \\
\hline 
ADM & \cmark & \xmark & & 10.94 & 6.02 & 100.98 & 0.69 & $\mathbf{0.63}$ \\
\base{} & \cmark & \cmark & 1.0 & 4.59 & 5.25 & 186.70 & 0.82 & 0.52 \\
\base{} & \cmark & \cmark & 10.0 & 9.11 & 10.93 & $\mathbf{283.92}$ & $\mathbf{0.88}$ & 0.32 \\
\our{} & \cmark & \cmark & 0.025 & $\mathbf{4.06}$ & $\mathbf{5.19}$ & 206.86 & 0.82 & 0.55 \\
\hline 
\end{tabular}
    \caption{Comparison between vanilla (\base{}~\cite{dhariwal2021diffusion}) and geometric (\our{}) guidance. \our{} produce better results across metrics when compared against \base{} variation optimized for highest FID scores. Evaluated on ImageNet 256x256 using 250 iterations during the sampling.}
    \label{tab:main_results}
\end{table*}

\paragraph{Definition of \our{}}
In the above reasoning, we have taken an arbitrary perturbation $v(x)$, which does not have to be consistent with the geometry of the data manifold $M$. Thus, if the perturbation (at least near $M$) is tangent to the manifold $M$, we can add a much larger perturbation and still not destroy the distance from the manifold given by \eqref{eq}. Observe that when we use the guidance in the backward process, we are close to the elements of the given class, and consequently, the gradient $\nabla p(y|x)$ of the classifier becomes tangent to the manifold $M$. Consequently, applying a larger constant than the baseline for classifier guidance does not lead to the unwanted behavior of leaving the predicted distance in \eqref{eq} from the manifold. Thus, we can take a function of $t$ which for large $t$ is similar to the previous $\sqrt{D(1-\bar\alpha_t)}/T$ (as for large $t$ we are far from the data manifold), while for small $t$ our trajectory is close to the manifold $M$ (the gradient of classifier becomes tangent to $M$) and we can choose higher values. Since $1-\bar \alpha_T\approx 1$, the simplest form of such strategy is given by\footnote{To compute the $\nabla p/\|\nabla p\|$ in a numerically stable way we use its equality to $\nabla \log p/\|\nabla \log p\|$}
$$
A_t=\frac{\sqrt{D}}{T} \frac{\nabla p(y|x)}{\|\nabla p(y|x)\|}, 
$$
which leads to \our{}, see Algorithm \ref{alg:our}.

\begin{figure*}
\centering
\begin{subfigure}{.33\textwidth}
  \centering
  \includegraphics[width=\linewidth]{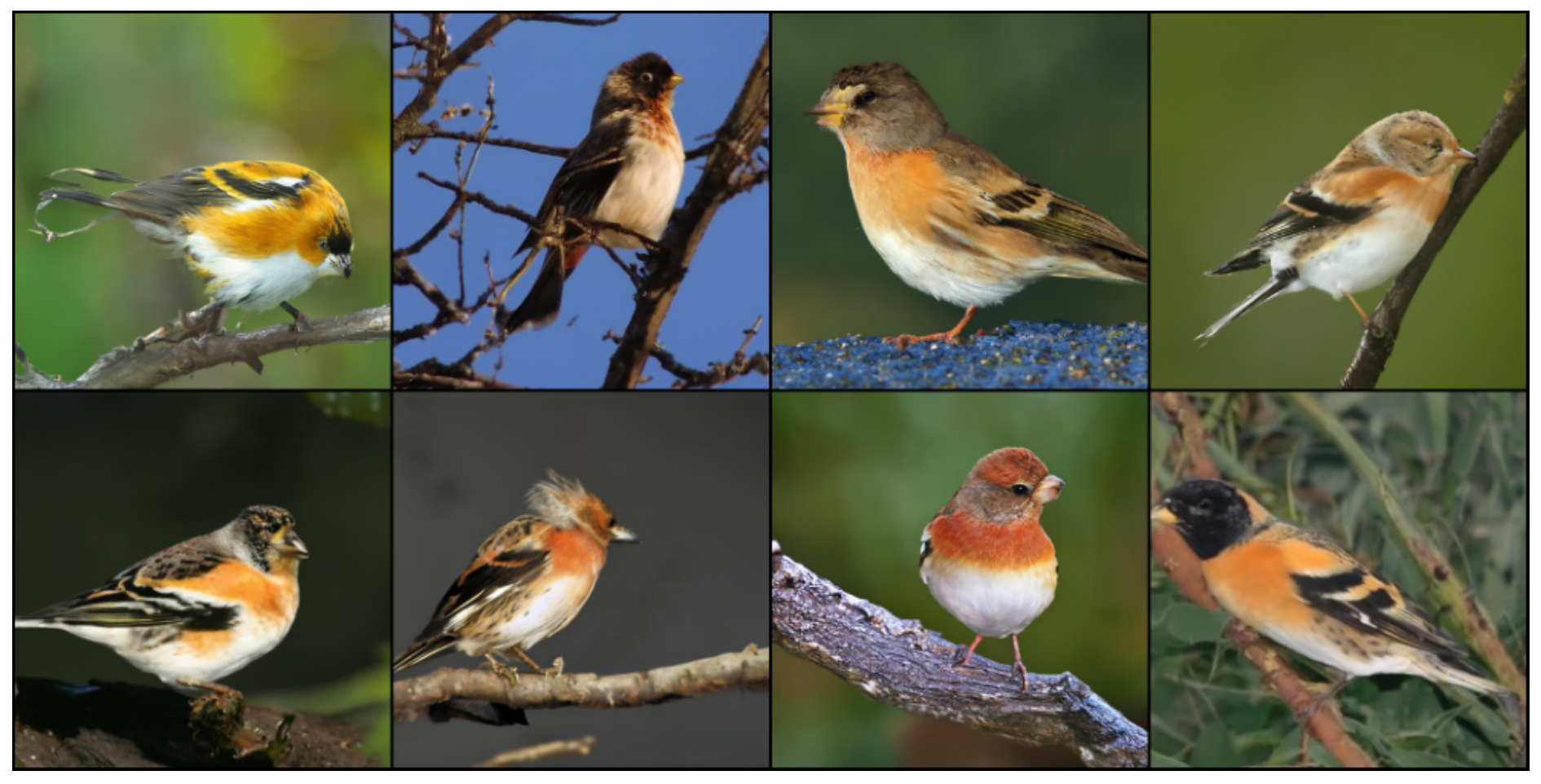}
  \includegraphics[width=\linewidth]{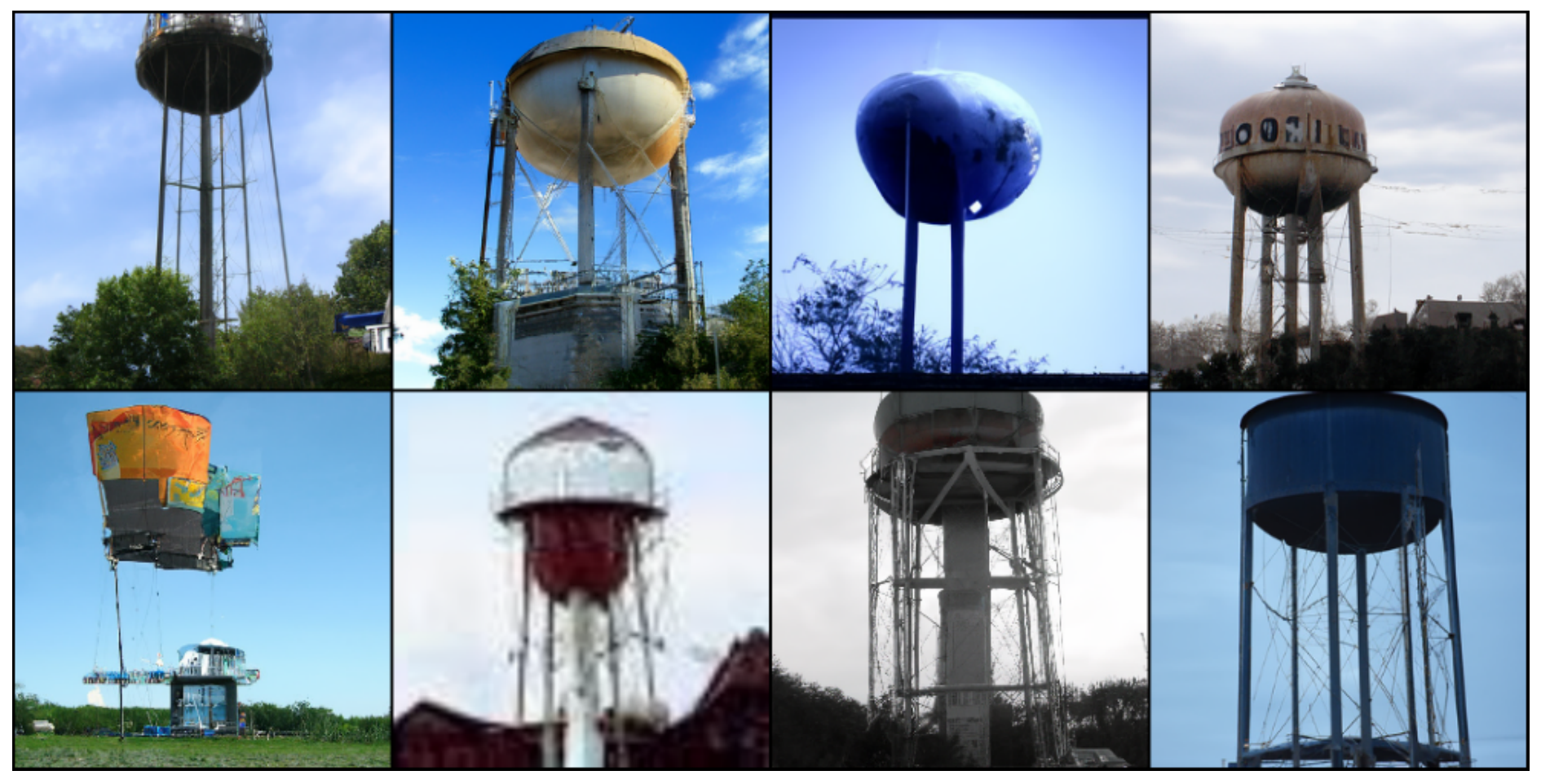}
  \includegraphics[width=\linewidth]{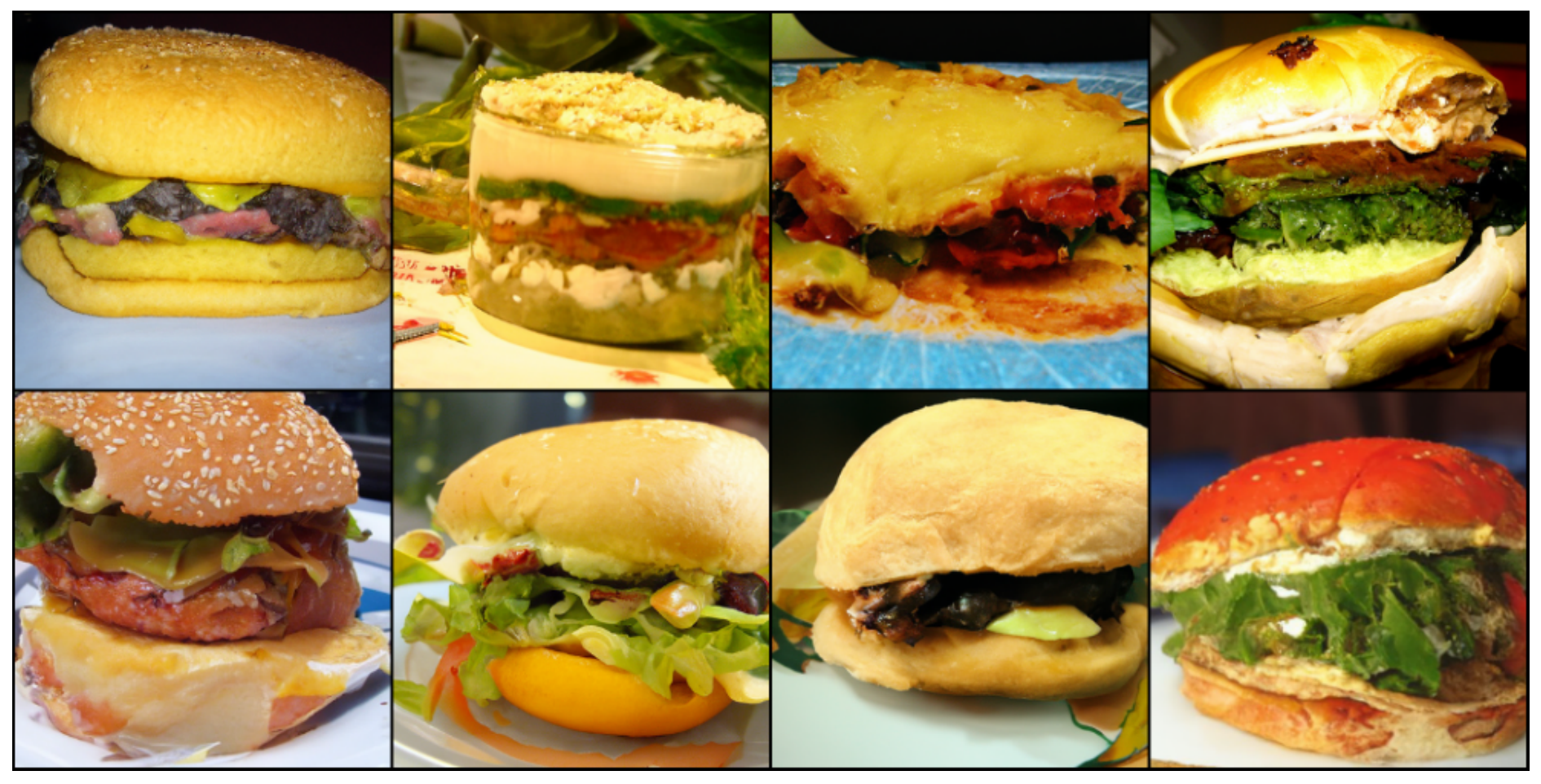}
  \caption{\base{}}
  \label{fig:base_distribution}
\end{subfigure}%
\begin{subfigure}{.33\textwidth}
  \centering
  \includegraphics[width=\linewidth]{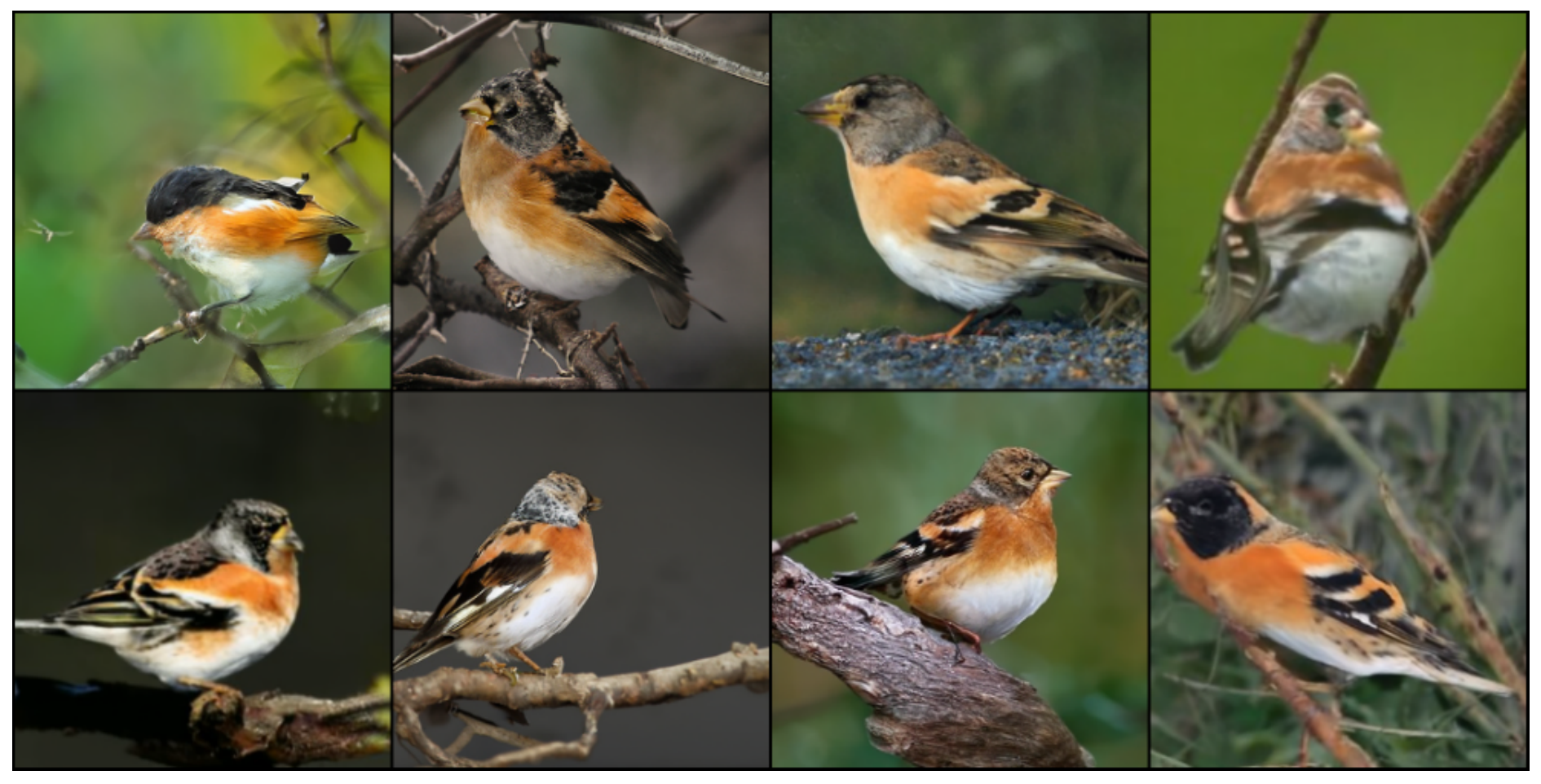}
  \includegraphics[width=\linewidth]{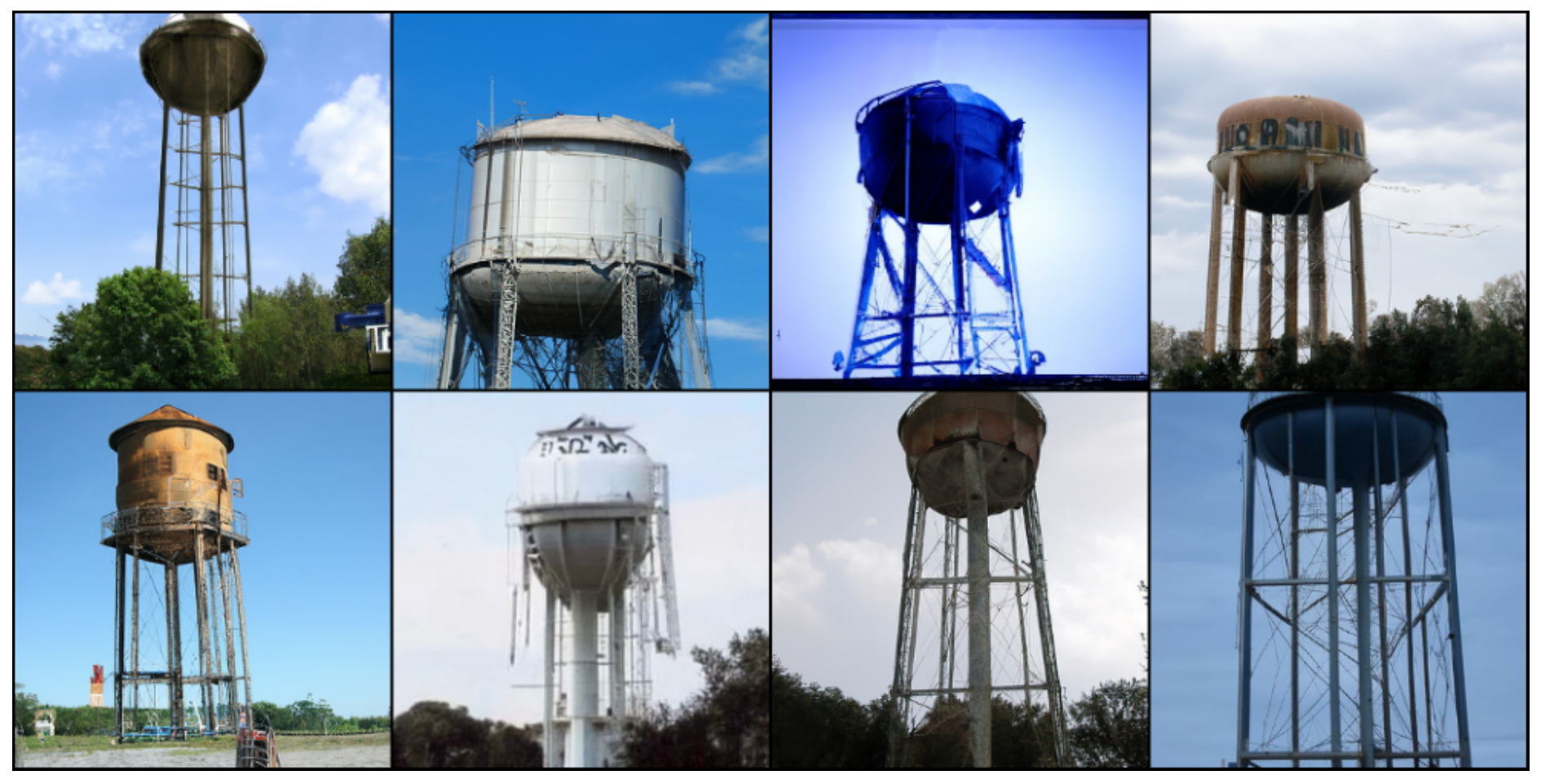}
  \includegraphics[width=\linewidth]{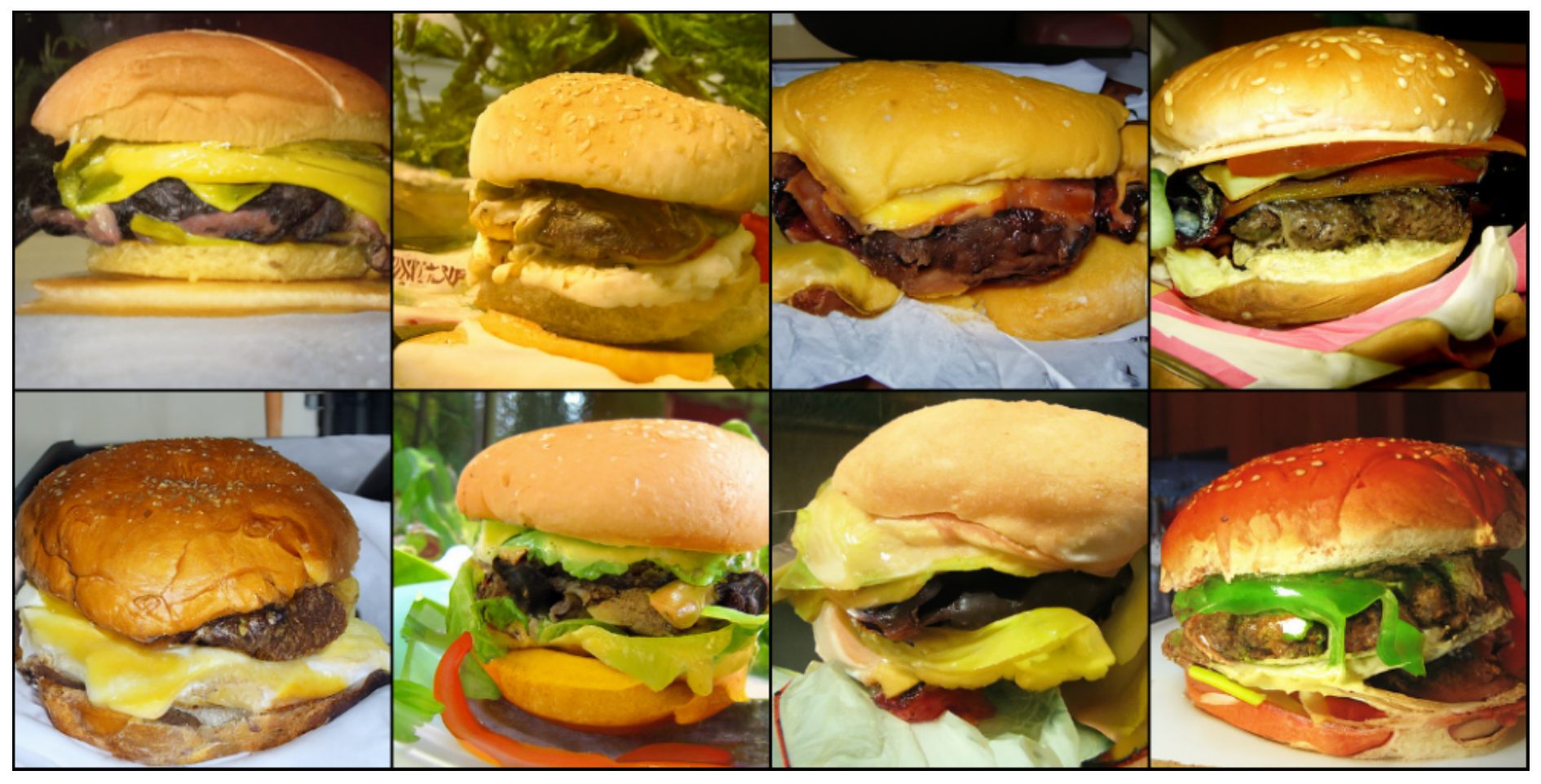}
  \caption{\our{}}
  \label{fig:our_distribution}
\end{subfigure}
\begin{subfigure}{.33\textwidth}
  \centering
  \includegraphics[width=\linewidth]{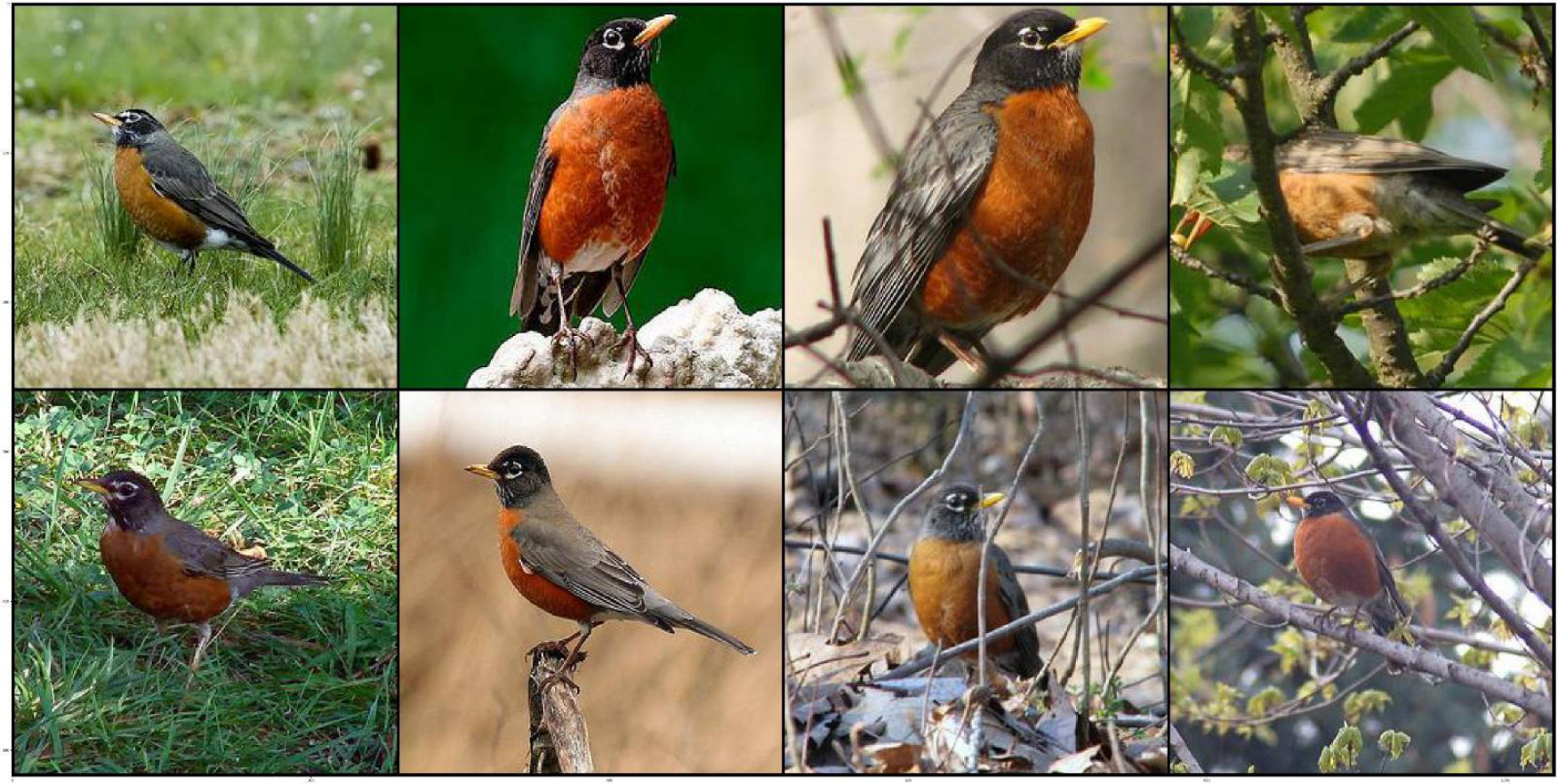}
  \includegraphics[width=\linewidth]{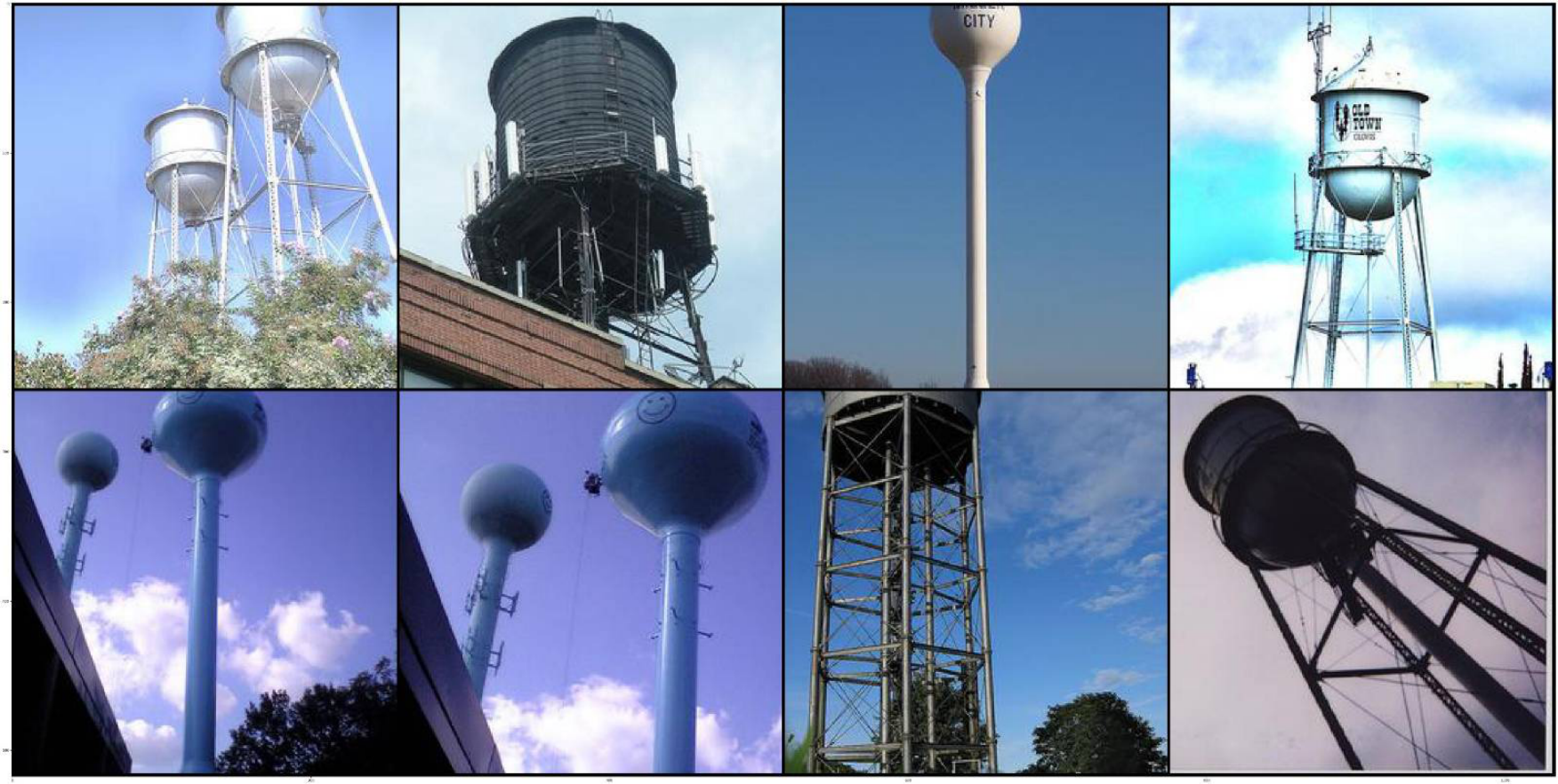}
  \includegraphics[width=\linewidth]{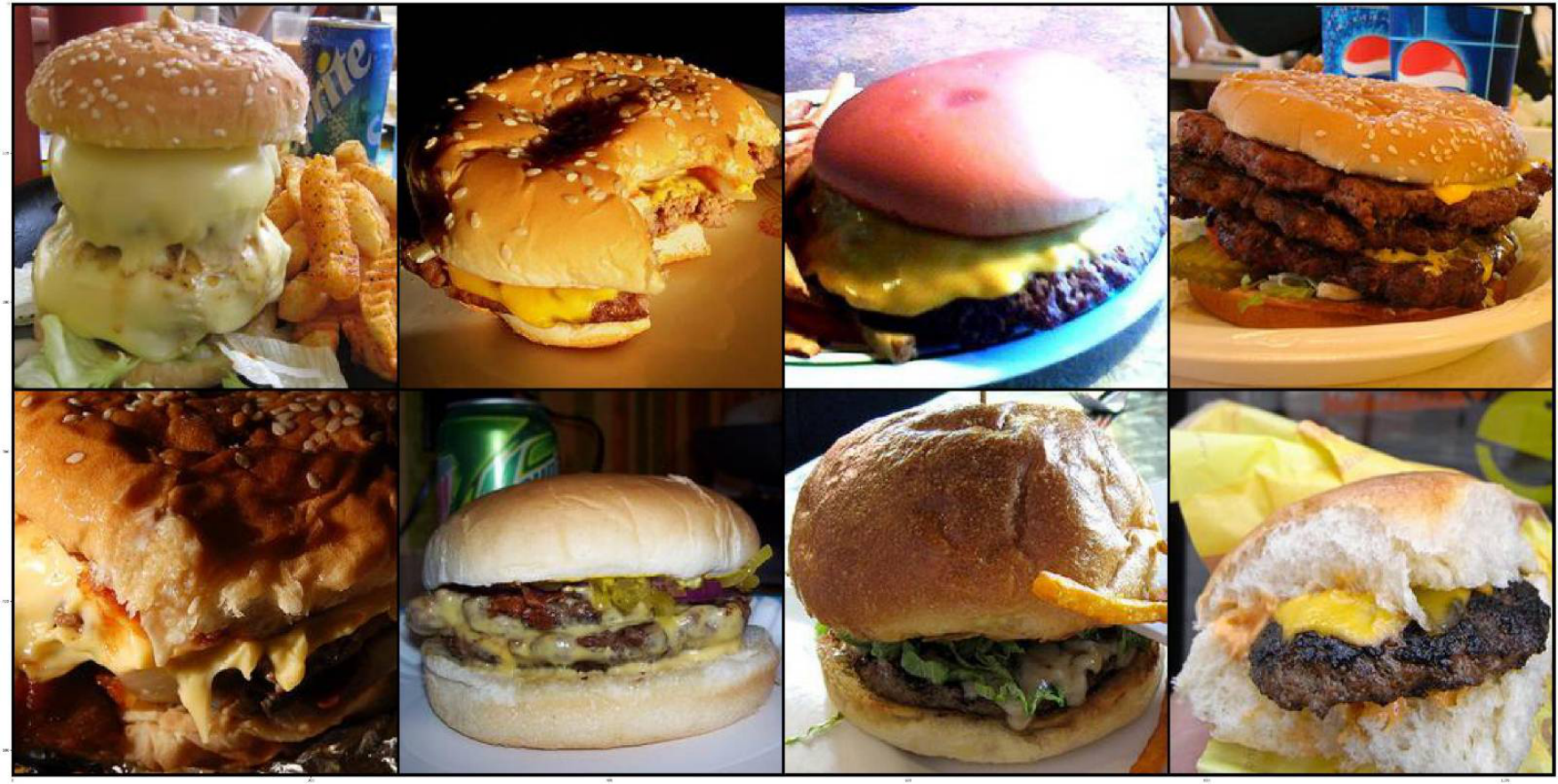}
  \caption{Dataset}
  \label{fig:dataset_distribution}
\end{subfigure}
\caption{Samples with vanilla classifier guidance \cite{dhariwal2021diffusion} (FID 12.00, left) vs samples with \our{} (FID 7.32, middle) and samples from the training set (right). Distribution of generated samples using both guidance methods is comparable, but significantly narrower compared to samples from original dataset. }
\label{fig:distribution_comparison}
\end{figure*}

\section{Experiments}

This section presents experiments that demonstrate the efficacy of the proposed method. The evaluation follows the protocol from \base{} \cite{dhariwal2021diffusion}. Additionally, a pre-trained model derived from the provided checkpoints and the authors' recommended sampling parameters are employed. Unless otherwise specified, experiments were conducted on ImageNet 256x256 images, sampled in 250 diffusion steps.

\begin{table}[]
\centering
\begin{tabular}{cccc}
\hline 
Model & Conditional & Scale & FID \\
\hline 
\our{} $(\sqrt{1-\bar \alpha_t})$ & \xmark & 0.15 & 7.47 \\
\our{} & \xmark & 0.15 & 7.32 \\
\hline 
\our{} $(\sqrt{1-\bar \alpha_t})$ & \cmark & 0.025 & 4.78 \\
\our{}  & \cmark & 0.025 & 4.06 \\
\hline 
\end{tabular}
    \caption{Comparison of \our{} with its variant given by \eqref{eq:var}, where we rescale the basic adjustment additionally by $\sqrt{1-\bar \alpha_t}$. Base approach achieves altogether better results. Evaluated on ImageNet 256x256 using 250 iterations during the sampling. }
    \label{tab:our_resuls_comparison}
    \vspace{-0.7cm}
\label{tab}
\end{table}

\paragraph{Quantitative comparison}
We used the same metrics as in \cite{dhariwal2021diffusion} to quantitatively evaluate our method. We pay the greatest attention to the values of the FID \cite{NIPS2017_8a1d6947} and use it as a primary metric to judge the results. It captures both the diversity and the fidelity of the generated samples and is widely used to compare generative models. We also used Inception Score (IS) \cite{salimans2016improvedtechniquestraininggans} and Precision \cite{kynkäänniemi2019improvedprecisionrecallmetric} to measure the fidelity of the images, as well as Recall \cite{kynkäänniemi2019improvedprecisionrecallmetric} to measure distribution coverage (diversity).

\begin{figure}
\centering
  \includegraphics[width=0.49\linewidth]{
  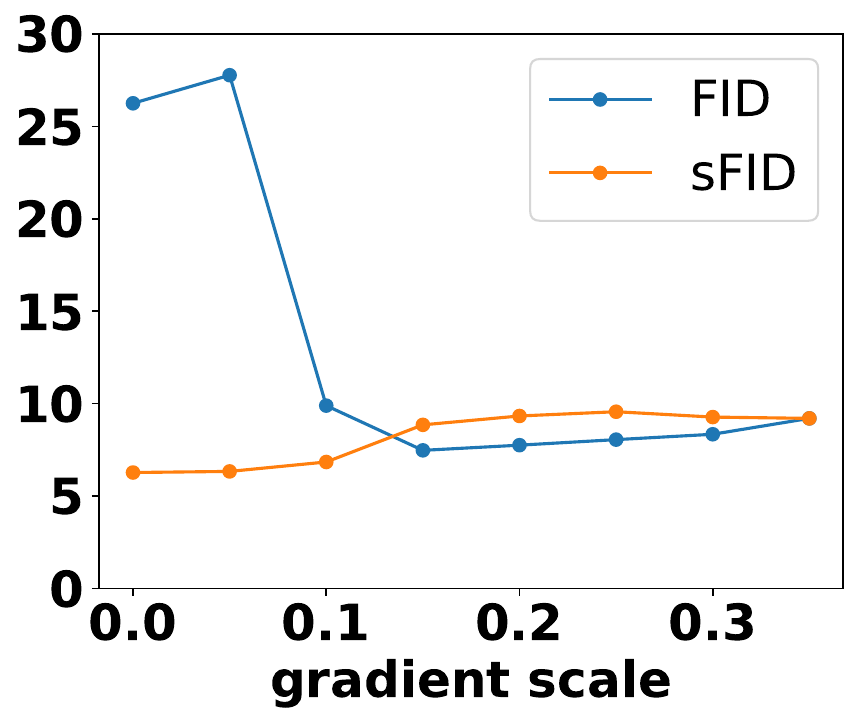}
  \includegraphics[width=0.49\linewidth]{
  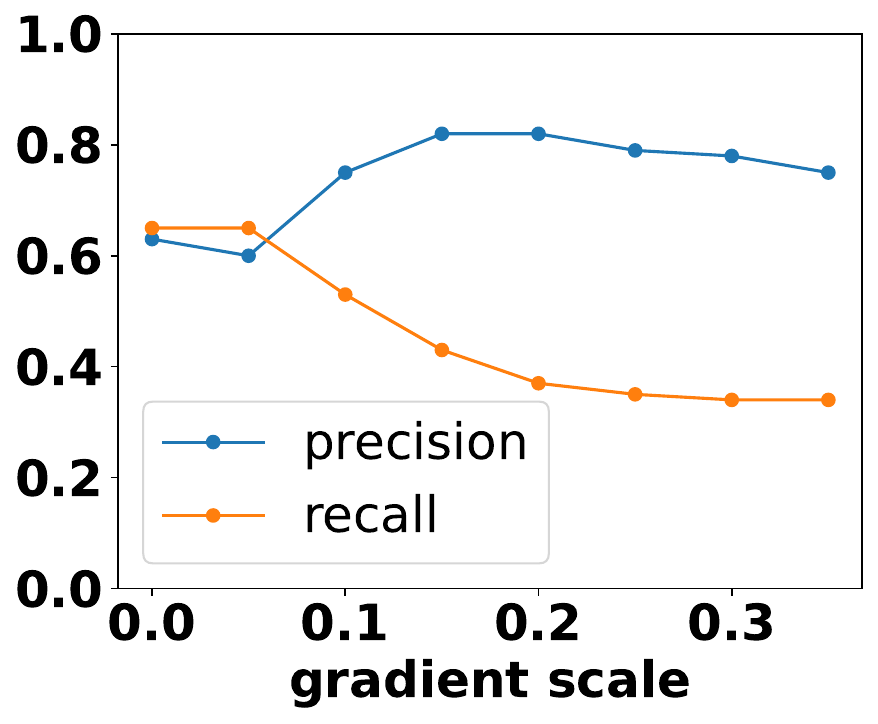}
\caption{Change in sample quality as we vary scale of the classifier gradients for unconditional ImageNet 256×256 model. It is possible to optimize for specific metrics by modifying the scale factor accordingly.}
\label{fig:quality_by_scale}
\end{figure}

In Table \ref{tab:main_results} we compare the results from \our{} with the vanilla classifier guidance (\base{})\cite{dhariwal2021diffusion}. To compute metrics, we generated 50000 random images and used a metric evaluation script with a reference batch provided by the authors. We are comparing models in conditional and unconditional setting.
We use a classifier scales $s=0.025$ and $s=0.15$ respectively, which turned out to be the best values for our approach to minimize the FID score. For unconditional setting, our method shows significant improvement (7.32 vs 12.00) in terms of FID values compared to the baseline approach. In the conditional case, the improvement is much smaller, but still significant (4.06 vs 4.59). Other metrics show improvement in our method as well, if we compare cases where the classifier scale parameter of the baseline model is also optimized for FID (see Figure~\ref{fig:quality_by_scale}). 

\begin{figure}[]
    \centering
    \includegraphics[width = 0.4\textwidth]{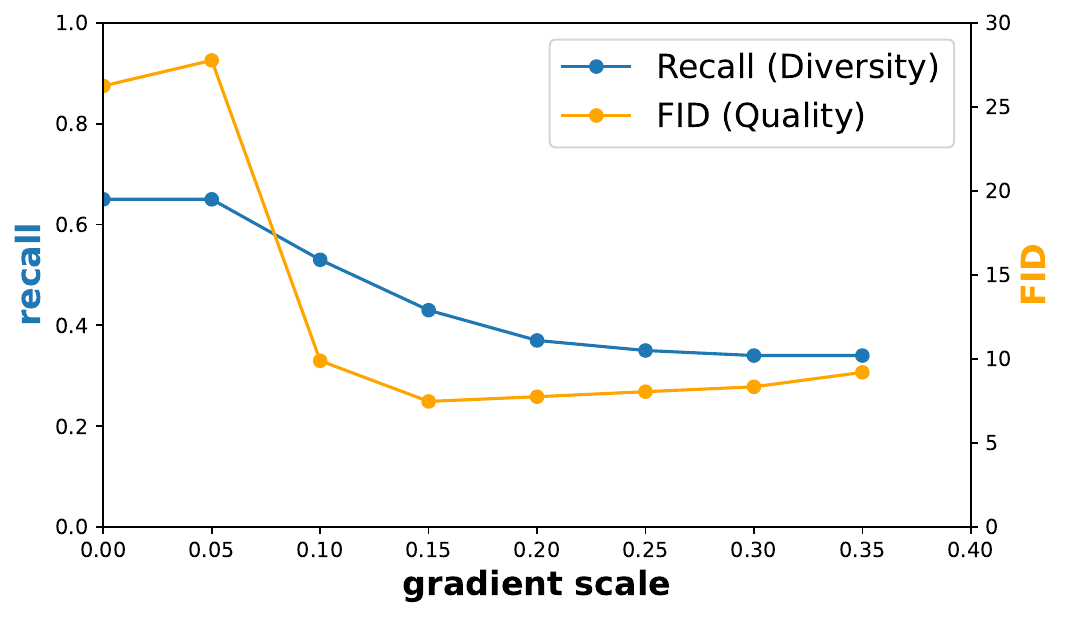}
    \caption{FID (Quality) and Recall (Diversity) trade-off in \our{}.}
    \label{fig:fid_vs_recall}
\end{figure}

The classifier guide in its base form introduces a trade-off between the quality and diversity of the generated images\cite{dhariwal2021diffusion}, measured, for example, by FID and Recall. In our approach, this relationship is also present. Figure \ref{fig:fid_vs_recall} shows how it changes depending on the varying classifier scale (guidance). We can observe that diversity (Recall) is the highest when we are not using guidance at all, and it is slowly decreasing as we strengthen the guidance. For the quality (FID) the relationship is reversed. 

\paragraph{Qualitative comparison}
In Figure \ref{fig:samples_comparison_uncond} we compare one-to-one samples generated with \base{} and \our{}. The unconditional model trained on ImageNet 256x256 was used. Each pair was sampled using the same starting and intermediary Gaussian noises and labels. We can observe that our samples are often sharper, with more details and class-specific features.

In Figure~\ref{fig:distribution_comparison} we compare the diversity of the generated images by looking at the entire batch of images of the same class. Ideally, images should be as varied as possible while still being classified in the proper class. Batches are created from the same random seed, the only difference being the guiding algorithm. We can notice that the distribution of generated samples using both guidance methods is comparable but significantly narrower compared to samples from the original data set. 


\paragraph{\our{} scaled} As part of our study, we experimented with a variation of \our{} motivated by \eqref{eq:var}, where we take $A_t = \sqrt{D}/T \sqrt{1-\bar \alpha_t} \nabla p/\|\nabla p\|$ instead of the baseline $A_t = \sqrt{D}/T \nabla p/\|\nabla p\|$. We denote such model by \our$(\sqrt{1-\bar \alpha_t})$. We thought that guiding toward a specific class is essential mainly at the early stages of the backward sampling process, when the image still forms from Gaussian noise. At later stages, it seemed like guidance should be progressively scaled down, as it would only provide irrelevant information about the class, which should already be encoded in the image itself. The numerical comparison of these two approaches is shown in Tab~\ref{tab:our_resuls_comparison}. As we can see, in the conditional and unconditional settings, the base approach is shown to produce better results.

\paragraph{Guidance cut-off}
As we observed in Figure~\ref{fig:modification-norm} guiding factor in \base{} quickly becomes close to 0. We think that this makes vanilla guidance effectively irrelevant for the majority of the sampling process, whereas \our{} can have a positive impact throughout the entire process. To observe this, we made an experiment where we turn off guidance after first 30\% of the sampling iterations. As we can see in Figure~\ref{fig:cut_guidance_base} for \base{} it didn't make a large difference in results, which means guidance in the following 70\% is not making a large impact. In \our{} we can see in Figure~\ref{fig:cut_guidance_our} that results are much worse with guidance cut-off, so guidance stays relevant also at later iterations.

\paragraph{\our{} combined with Robust Classifier}\label{rob_classifier}

As \our{} can be easily incorporated into existing models, we tried to combine it with robust classifier~\cite{kawar2023enhancingdiffusionbasedimagesynthesis}, which was previously mentioned in Section~\ref{rob_classifier_descr}. In Table~\ref{tab:robust_compare} we can see that using both of these methods together further improves sampling quality and achieves the the best results in terms of quality metrics. Unfortunately, we didn't have pre-trained weights required to compare them  in unconditional setting, where we would expect that results would be even more prominent.  

\begin{table}[]
\centering
\begin{tabular}{@{}c@{\;}c@{\;}c@{\;}c@{\;}c@{\;}c@{}}
\hline 
Model & FID & sFID & IS & Precision & Recall \\
\hline 
\base{} & 2.97 & 5.09 & - & 0.78 & 0.59 \\
Robust & 2.85 & - & - & 0.82 & 0.56 \\
\our{} & 2.83 & 5.17 & 151.63 & 0.80 & 0.61 \\
\our{} + Robust & 2.81 & 5.16 & 152.37 & 0.80 & 0.60 \\
\hline 
\end{tabular}
    \caption{Combining \our{} with robust classifier on ImageNet 128x128 using a conditional model. \our{} improves results independently and with a robust classifier. Missing sFID and IS values were not present in the original papers.}
    \label{tab:robust_compare}
\end{table}

\section{Conclusion}

This paper proposes \our{}, a method for guiding diffusion models following the distance between the denoising trajectory and the data manifold. Our metric approach can use similar guidance during the entire denoising process to obtain sharper images. In classical methods, such guidance is more critical at the beginning of the denoising process.  
\our{} is easy to execute because it depends on classifier gradient normalization and outperforms the probabilistic method ADM-G regarding FID scores and the quality of the images produced.


\section{Comparison of samples from conditional model }

\begin{figure*}[h!]
    \centering
    \begin{tabular}{@{}c@{}c@{}}
        \rotatebox{90}{ \, \quad \our{} \qquad \base{}} &
        \includegraphics[width = 0.9\textwidth]{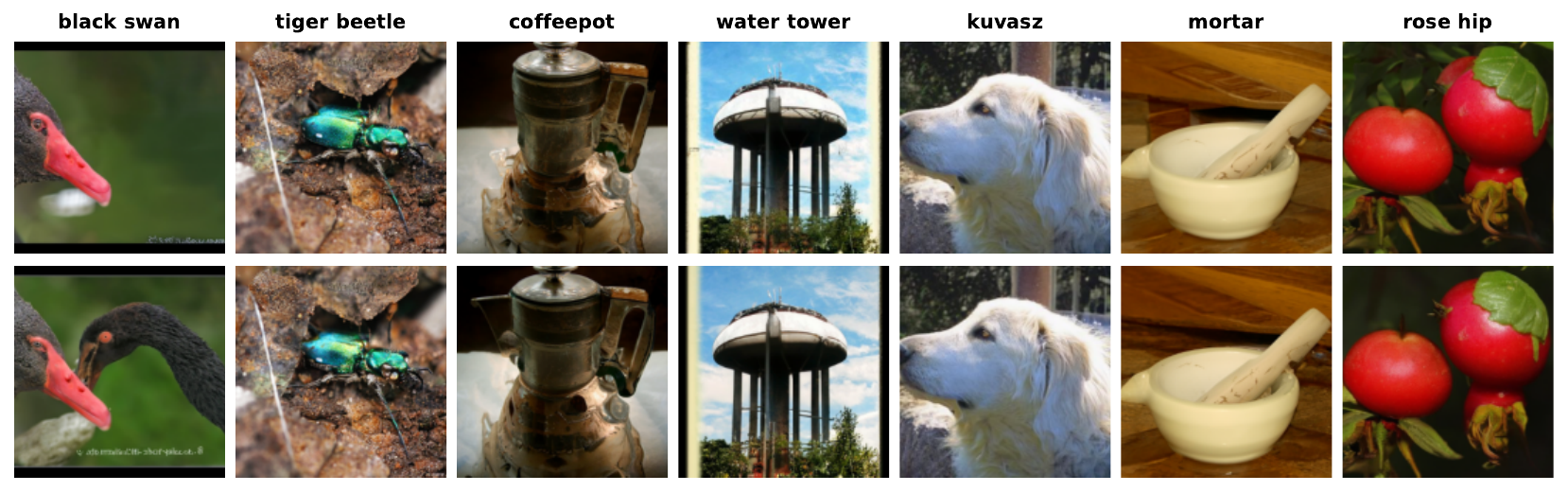} \\
        \rotatebox{90}{ \, \quad \our{} \qquad \base{}} &
        \includegraphics[width = 0.9\textwidth]{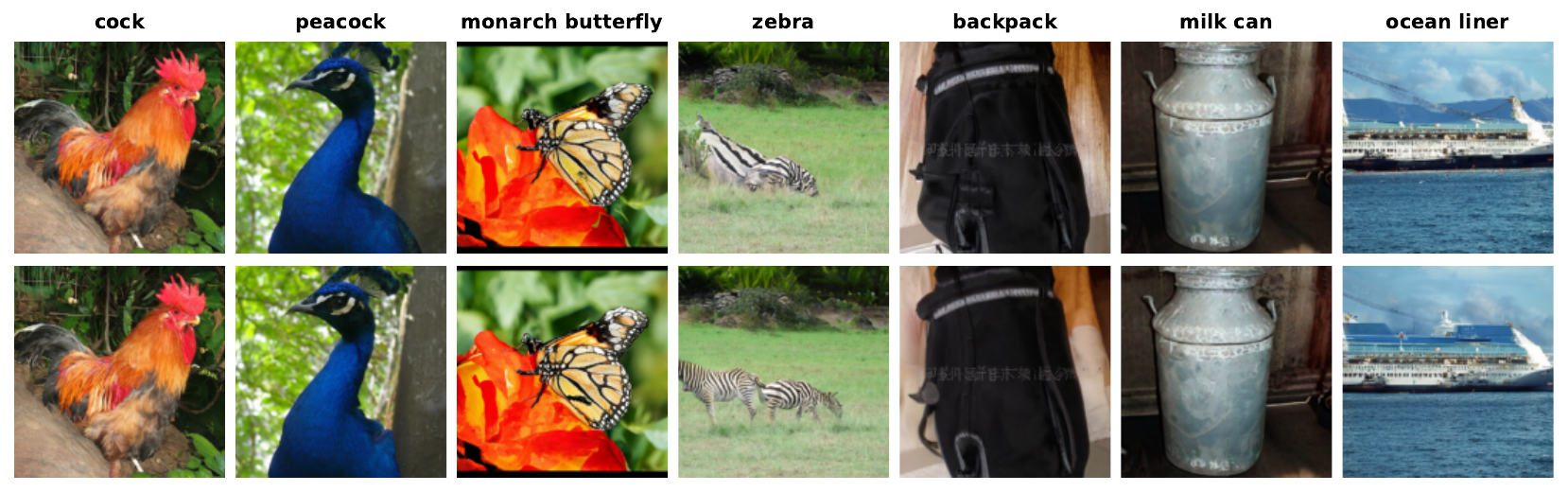} \\
        \rotatebox{90}{ \, \quad \our{} \qquad \base{}} &
        \includegraphics[width = 0.9\textwidth]{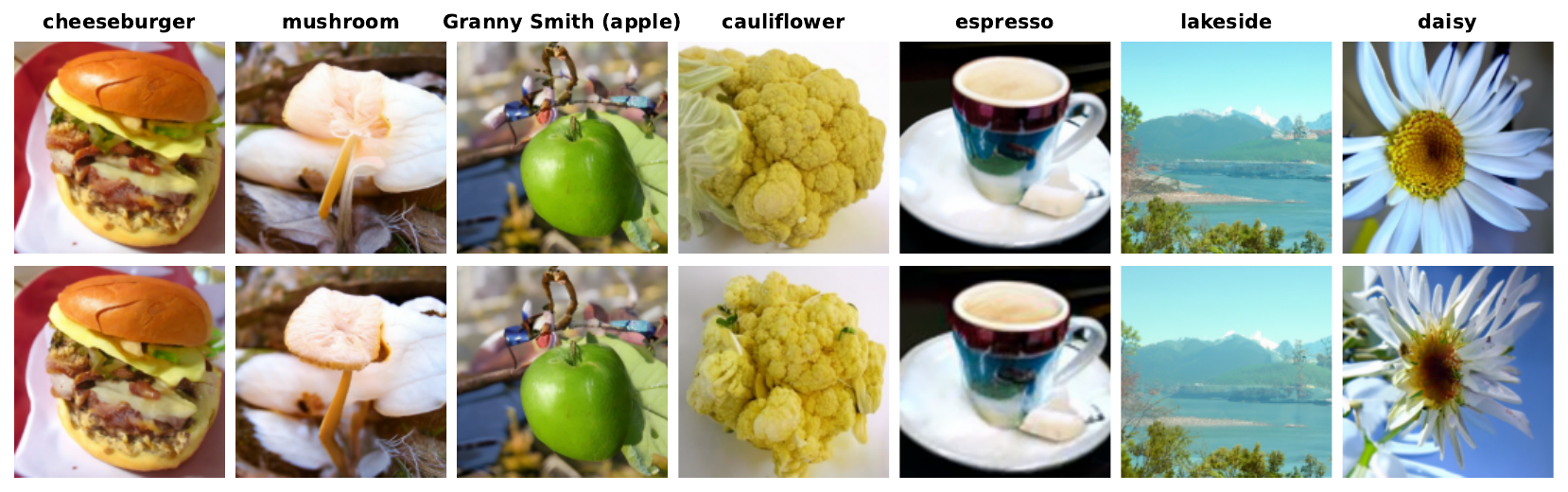}\\
    \end{tabular}
    \caption{Images generated by conditional guided diffusion using the same noise (random seed) and class label, with a vanilla (FID 4.78, top) and a geometric (FID 4.06, bottom) guidance. Visually both models produce mostly similar results, possibly with small improvement in \our{}. Difference in quality is much smaller compared with unconditional model. }
    \label{fig:samples_comparison_cond_alt}
\end{figure*}

In Figure~\ref{fig:samples_comparison_cond_alt} we compare one-to-one samples generated with \base{} and \our{}. Conditional model trained on ImageNet 256x256 was used. Each pair was sampled using the same random seed and class labels. We can observe that visually improvement is much smaller compared to unconditional case, as models produce mostly similar results.

\section{Examination of the intermediate denoising steps}

\begin{figure*}[h!]
    \centering
    \begin{tabular}{@{}c@{}c@{}}
        \rotatebox{90}{ \small \our{} \quad ADM-G} &
        \includegraphics[width = 0.9\textwidth]{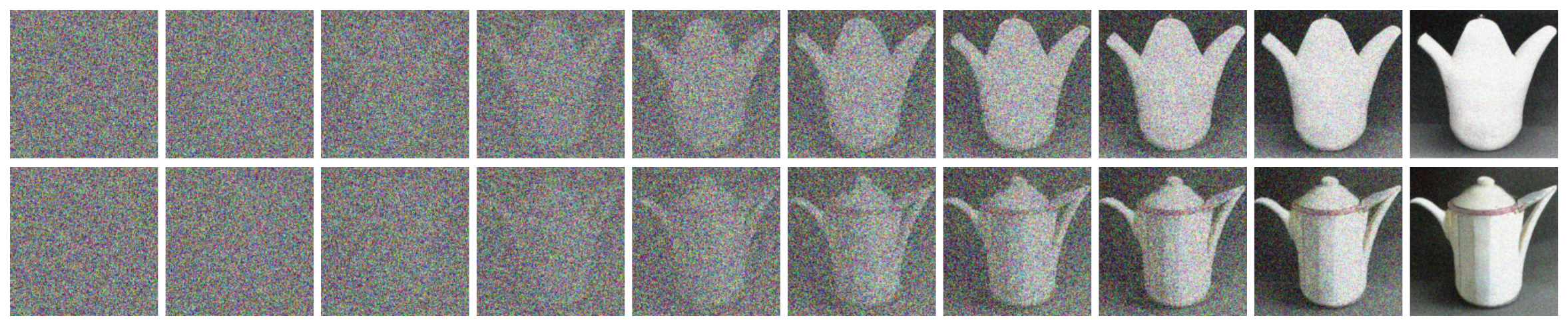} \\
        \rotatebox{90}{ \small \our{} \quad ADM-G} &
        \includegraphics[width = 0.9\textwidth]{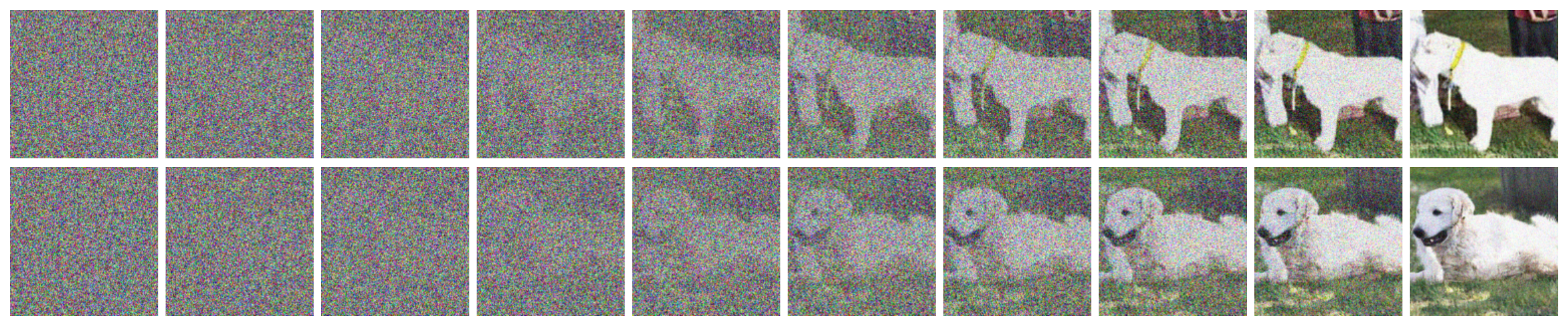} \\
        \rotatebox{90}{ \small \our{} \quad ADM-G} &
        \includegraphics[width = 0.9\textwidth]{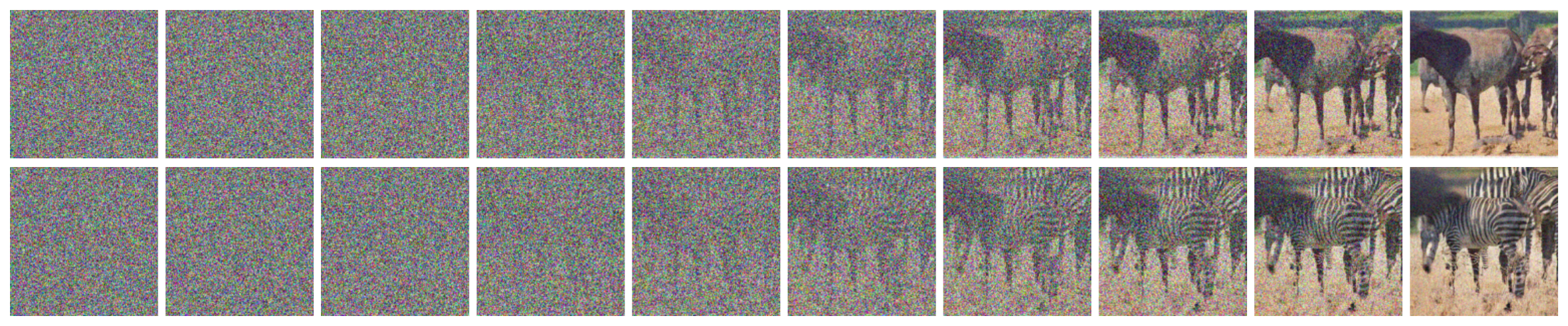}\\
        \rotatebox{90}{ \small \our{} \quad ADM-G} &
        \includegraphics[width = 0.9\textwidth]{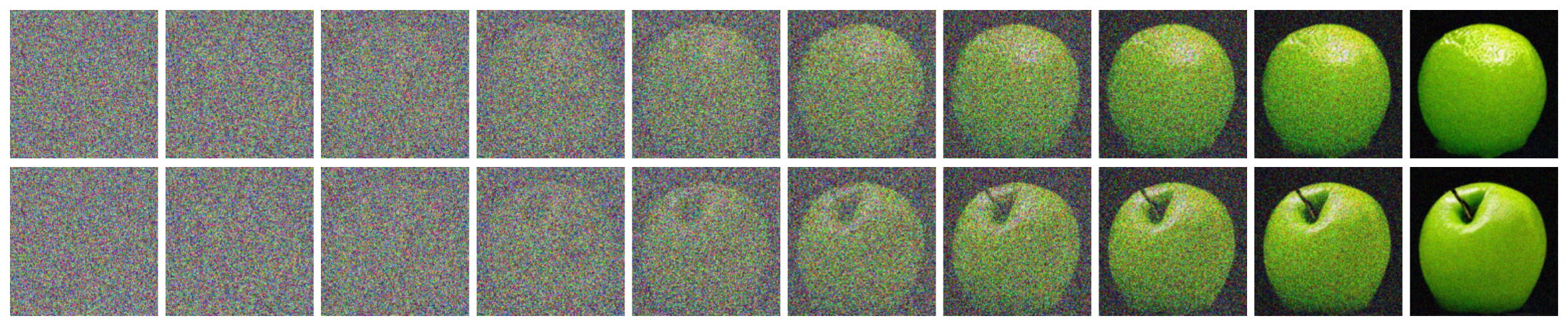}\\
    \end{tabular}
    \caption{Comparison of the selected snapshots from the denoising processes. \our{} impoves quality over the entire timeline. } 
    \label{fig:samples_intermediate_steps}
\end{figure*}

Figure~\ref{fig:samples_intermediate_steps} compares corresponding intermediate steps during the sampling process with \our{} and \base{}. The same class label and random seed was used for both models. We can observe that samples from \our{} are visually more appealing even at earlier stages of the process, which can suggest that \our{} brings improvement over vanilla guidance during the entire process, not only at the later stages.

\section{Examining number of forward passes in \our{}}

Diffusion models during the training usually uses 1000 diffusion steps, which can be later reduced during the sampling stage. This can significantly accelerate the sampling speed, without making a large sacrifice in quality. Table \ref{tab:sampling_iters_comparison} shows this relation in case of \our{}. We can observe similar results for all tested settings. Using 1000 iterations brings very little improvement over 250, and even using just 50 steps can produce satisfying results if we want to significantly reduce sampling time. 

\begin{table*}[h!]
\centering
\begin{tabular}{ccccccccc}
\hline 
Model & Conditional & Iters & Scale & FID & sFID & IS & Precision & Recall \\
\hline 
\our{} $(\sqrt{1-\bar \alpha_t})$ & \xmark & 50 & 0.15 & 9.61 & 13.45 & 185.63 & 0.78 & 0.41 \\
\our{} $(\sqrt{1-\bar \alpha_t})$& \xmark & 250 & 0.15 & 7.47 & 8.85 & 206.79 & 0.82 & 0.43 \\
\our{} $(\sqrt{1-\bar \alpha_t})$ & \xmark & 1000 & 0.15 & 7.15 & 7.73 & 208.33 & 0.82 & 0.43 \\
\hline 
\our{}  & \xmark & 50 & 0.15 & 9.80 & 11.41 & 226.90 & 0.72 & 0.40 \\
\our{}  & \xmark & 250 & 0.15 & 7.32 & 7.98 & 243.34 & 0.77 & 0.42 \\
\our{}  & \xmark & 1000 & 0.15 & 7.14 & 7.43 & 245.79 & 0.78 & 0.42 \\
\hline 
\our{}  & \cmark & 50 & 0.025 & 6.10 & 8.53 & 191.41 & 0.79 & 0.52 \\
\our{}  & \cmark & 250 & 0.025 & 4.06 & 5.19 & 206.86 & 0.82 & 0.55 \\
\our{}  & \cmark & 1000 & 0.025 & 4.16 & 4.98 & 202.65 & 0.83 & 0.55 \\
\hline 
\end{tabular}
    \caption{Effect of using different number of forward passes during the sampling process with \our{} and its variation on ImageNet 256x256. Using 1000 iterations is not providing significant improvement over 250, in contrast to using 50, where difference is much larger. Taking into account computation time, using 250 iterations seems optimal for most use-cases. }
    \label{tab:sampling_iters_comparison}
\end{table*}

\section{\our{} scaled variant}
Table~\ref{tab:our_resuls_comparison} is a full version of the table with comparison of \our{} with its variant given by $A_t=\frac{\sqrt{D}}{T}\sqrt{1-\bar \alpha_t} \frac{v(x)}{\|v(x)\|}$.

\begin{figure*}[]
    \centering
    \includegraphics[width = 0.95\textwidth]{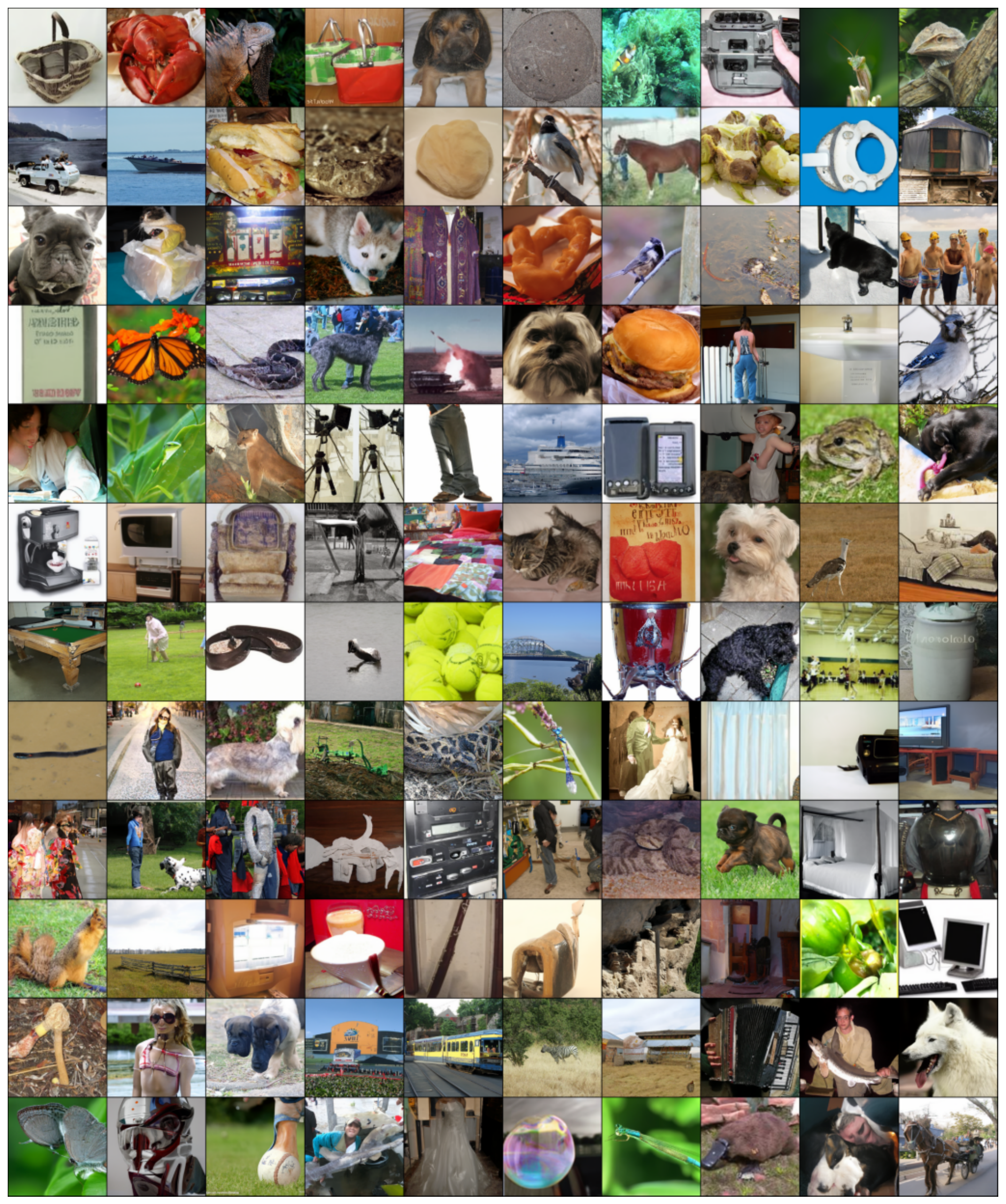} \\
    \caption{Uncurated samples from ImageNet 256x256 generated by a conditional diffusion model, guided with \our{} (FID 4.06). 
    }
    \label{fig:samples_random_cond}
\end{figure*}

\begin{figure*}[h!]
    \centering
    \includegraphics[width = 0.95\textwidth]{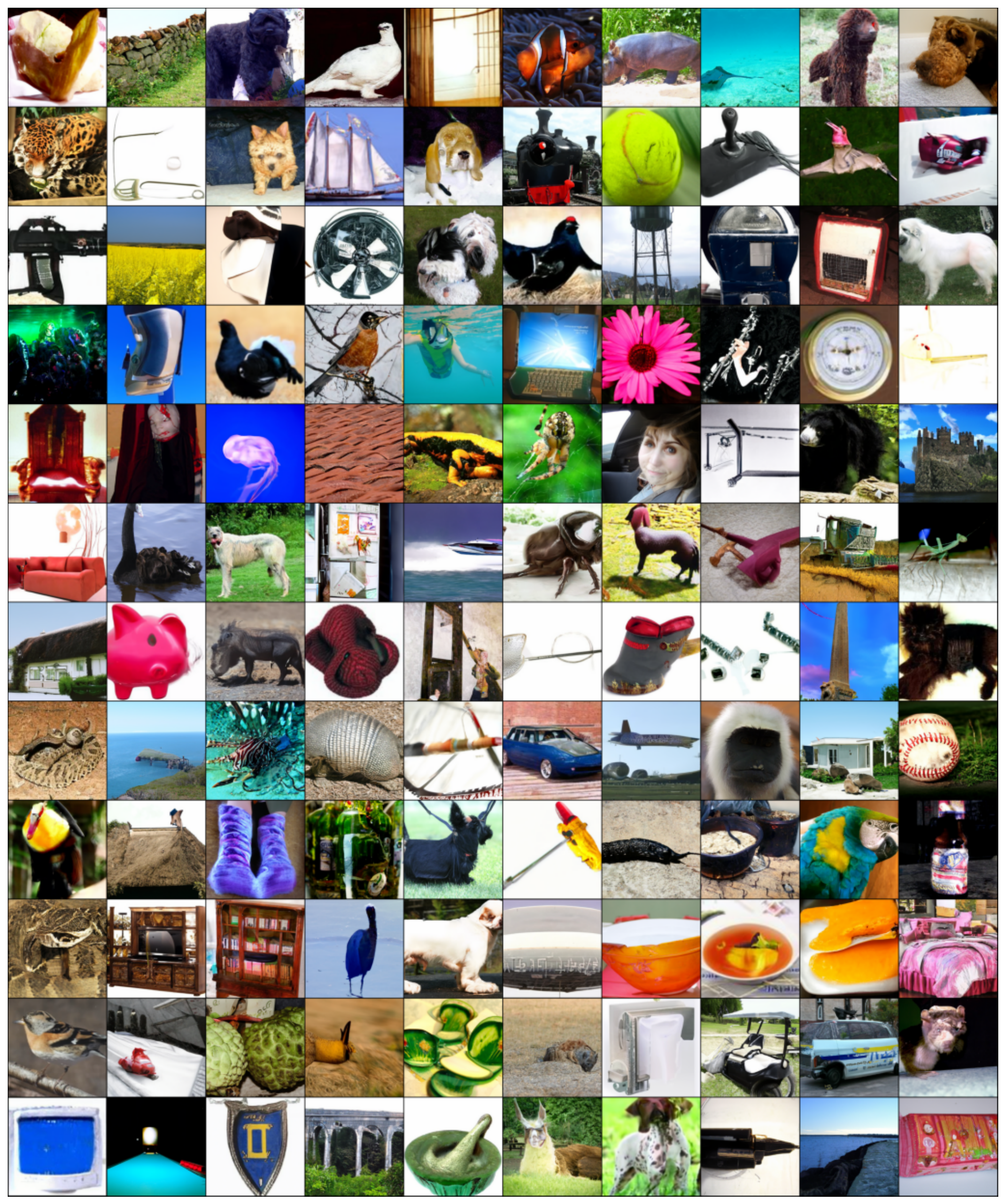} \\
    \caption{Uncurated samples from ImageNet 256x256 generated by a unconditional diffusion model, guided with \our{} (FID 7.32). 
    }
    \label{fig:samples_random_uncond}
\end{figure*}

\begin{table*}[h!]
\centering
\begin{tabular}{cccccccc}
\hline 
Model & Conditional & Scale & FID & sFID & IS & Precision & Recall \\
\hline 
\our{} $(\sqrt{1-\bar \alpha_t})$ & \xmark & 0.15 & 7.47 & 8.85 & 206.79 & 0.82 & 0.43 \\
\our{} & \xmark & 0.15 & 7.32 & 7.98 & 243.34 & 0.77 & 0.42 \\
\hline 
\our{} $(\sqrt{1-\bar \alpha_t})$ & \cmark & 0.025 & 4.78 & 5.13 & 174.13 & 0.80 & 0.56 \\
\our{}  & \cmark & 0.025 & 4.06 & 5.19 & 206.86 & 0.82 & 0.55 \\
\hline 
\end{tabular}
    \caption{Comparison of \our{} with its variant, where we rescale the basic adjustment additionally by $\sqrt{1-\bar \alpha_t}$. Base approach achieves altogether better results. Evaluated on ImageNet 256x256 using 250 iterations during the sampling. }
    \label{tab:our_resuls_comparison}
\end{table*}

\end{document}